%% file: main.tex
\documentclass[runningheads]{llncs}
\usepackage{graphicx}
\usepackage{comment}
\usepackage{amsmath,amssymb} % define this before the line numbering.
\usepackage{color}

\usepackage{cite}
\usepackage{epsfig}
\usepackage{graphicx}
\usepackage{amsmath}
\usepackage{amssymb}
\usepackage{multirow}
\usepackage{booktabs}

\usepackage[normalem]{ulem}
\usepackage{ulem}

\usepackage[pagebackref=true,breaklinks=true,letterpaper=true,colorlinks,bookmarks=false]{hyperref}

\begin{document}
\setlength\floatsep{4truemm}
\setlength\textfloatsep{4truemm}
\setlength\abovecaptionskip{1truemm}
% \linenumbers
\pagestyle{headings}
\mainmatter

\title{Do We Need Sound for Sound Source Localization?} % Replace with your title

\titlerunning{Do We Need Sound for Sound Source Localization?}
\authorrunning{T. Oya et al.}
\author{Takashi Oya* \and Shohei Iwase* \and Ryota Natsume \and Takahiro Itazuri \and \\ Shugo Yamaguchi \and Shigeo Morishima}
\institute{Graduate School of Advanced Science and Engineering,\\Waseda University, Tokyo, Japan}
%\begin{comment}
%\titlerunning{ACCV-20 submission ID \ACCV20SubNumber}
%\authorrunning{ACCV-20 submission ID \ACCV20SubNumber} 
%\author{Anonymous ACCV 2020 submission}
%\institute{Paper ID \ACCV20SubNumber}
%\end{comment}
%******************

\begin{comment}
\titlerunning{Abbreviated paper title}
\author{First Author\inst{1}\orcidID{0000-1111-2222-3333} \and
Second Author\inst{2,3}\orcidID{1111-2222-3333-4444} \and
Third Author\inst{3}\orcidID{2222--3333-4444-5555}}
%
\authorrunning{F. Author et al.}
\institute{Princeton University, Princeton NJ 08544, USA \and
Springer Heidelberg, Tiergartenstr. 17, 69121 Heidelberg, Germany
\email{lncs@springer.com}\\
\url{http://www.springer.com/gp/computer-science/lncs} \and
ABC Institute, Rupert-Karls-University Heidelberg, Heidelberg, Germany\\
\email{\{abc,lncs\}@uni-heidelberg.de}}
\end{comment}
%******************
\maketitle

%%%%%%%%% ABSTRACT
\begin{abstract}
\input{sec_00_abstract}
\end{abstract}

%%%%%%%%% BODY TEXT
\input{sec_01_introduction}
\input{sec_02_related_work}

\input{sec_03_proposed_framework}
\input{sec_04_experiments}

\input{sec_05_improved_eval}
\input{sec_06_conclusion}

\bibliographystyle{splncs}
\bibliography{egbib}
\clearpage
\newpage

\appendix

\title{Supplementary Material: \\Do We Need Sound for Sound Source Localization?} % Replace with your title

%\begin{comment}
\titlerunning{Do We Need Sound for Sound Source Localization?}
\authorrunning{T. Oya et al.}
\author{Takashi Oya* \and Shohei Iwase* \and Ryota Natsume \and Takahiro Itazuri \and \\ Shugo Yamaguchi \and Shigeo Morishima}
\institute{Graduate School of Advanced Science and Engineering,\\Waseda University, Tokyo, Japan}
%\end{comment}
%******************

\begin{comment}
\titlerunning{Abbreviated paper title}
\author{First Author\inst{1}\orcidID{0000-1111-2222-3333} \and
Second Author\inst{2,3}\orcidID{1111-2222-3333-4444} \and
Third Author\inst{3}\orcidID{2222--3333-4444-5555}}
%
\authorrunning{F. Author et al.}
\institute{Princeton University, Princeton NJ 08544, USA \and
Springer Heidelberg, Tiergartenstr. 17, 69121 Heidelberg, Germany
\email{lncs@springer.com}\\
\url{http://www.springer.com/gp/computer-science/lncs} \and
ABC Institute, Rupert-Karls-University Heidelberg, Heidelberg, Germany\\
\email{\{abc,lncs\}@uni-heidelberg.de}}
\end{comment}
%******************
\maketitle

%%%%%%%%% ABSTRACT

%%%%%%%%% BODY TEXT
\input{supplement}
\end{document}

%% file: sec_00_abstract.tex
During the performance of sound source localization which uses both visual and aural information, it presently remains unclear how much either image or sound modalities contribute to the result, i.e. do we need both image and sound for sound source localization?
To address this question, we develop an unsupervised learning system that solves sound source localization by decomposing this task into two steps: 
(i) ``potential sound source localization'', a step that localizes possible sound sources using only visual information
(ii) ``object selection'', a step that identifies which objects are actually sounding using aural information.
Our overall system achieves state-of-the-art performance in sound source localization, and more importantly, we find that despite the constraint on available information, the results of (i) achieve similar performance.
From this observation and further experiments, we show that visual information is dominant in ``sound'' source localization when evaluated with the currently adopted benchmark dataset. Moreover, we show that the majority of sound-producing objects within the samples in this dataset can be inherently identified using only visual information, and thus that the dataset is inadequate to evaluate a system's capability to leverage aural information. As an alternative, we present an evaluation protocol that enforces both visual and aural information to be leveraged, and verify this property through several experiments. 

\keywords{cross-modal learning, sound source localization, unsupervised learning}

%% file: sec_01_introduction.tex
\input{fig_concept.tex}

\section{Introduction}
In many scientific areas, it is common to make a prediction of an objective variable using explanatory variables.  Nevertheless, the purpose is not always to generate an accurate prediction, but to investigate which variable is important for a prediction.  For example, when using learning methods such as Random Forests \cite{randomforest}, or their variants like Gradient Boosted Trees \cite{xgboost, lightgbm, catboost}, quantities such as ``variable importance'' are widely used to determine the predictive strength of specific variables. In a similar manner, linear regression coefficients are known to be very useful to evaluate the contribution of specific variables in a linear regression analysis. However, there are only a few studies that investigate the contributions of each modality in cross-modal perception tasks that leverage both visual and aural information. Our study focuses on the investigation of this issue in sound source localization, which is a cross-modal perception task aiming to determine image pixels that are associated with a given sound source, e.g. selecting the region of trumpet in an image when given sound of trumpet. 

In this research, we develop an unsupervised architecture which solves sound source localization by decomposing the problem into two tasks. First, given only an image, our system initially suggests a potential localization map, which is a heat map of all objects that are capable of producing sound.  We call this step ``potential sound source localization''.  Then, given a sound, our system identifies which objects within the potential localization map are actually producing sound, and returns a localization map. By comparing the performance of the potential localization map, which is derived only from visual information, and the localization map, which leverages both visual and aural information, we are able to evaluate the contributions of each modality in the benchmark dataset. 

As can be seen from Fig. \ref{fig:concept}, the architecture has 3 components: the potential localization network, the sound network, and the selection module. The potential localization network is an image network that takes an image as an input and returns the potential localization map. Notably, the potential localization network operates independently from the sound network, which returns latent features derived from the sound input. The selection module updates the potential localization map to the localization map using the latent features, by selecting objects from the potential localization map. During the training phase, our model can be trained in an unsupervised manner as we utilize audio-visual correspondence \cite{LLL}, i.e. whether image and sound correspond to each other or not. 

As a result, when using both visual and aural information, our system achieves state of the art performance. More significantly, although utilizing only visual information, the potential localization map performs similarly to the localization map, indicating that in the current benchmark dataset, high performance can be achieved without the usage of sound as input.
Furthermore, from additional quantitative and qualitative analysis, we have found that for the majority of the samples within the current dataset, the sound-producing object can be inherently identified using only image, as all objects that are capable of producing sound are actually producing sound. This is problematic as this is not the case in the real world, and proper evaluation on a system's capability of leveraging aural information can not be conducted. To overcome this issue, we design an evaluation protocol called {\it concat-and-localize}, where evaluation is conducted on artificially created samples that enforce the usage of aural information.  We verify this property through evaluation of our system using this evaluation protocol. 

%% file: fig_concept.tex
\begin{figure}[t]
    \centering
    \scalebox{0.8}{
        \includegraphics[width=\linewidth]{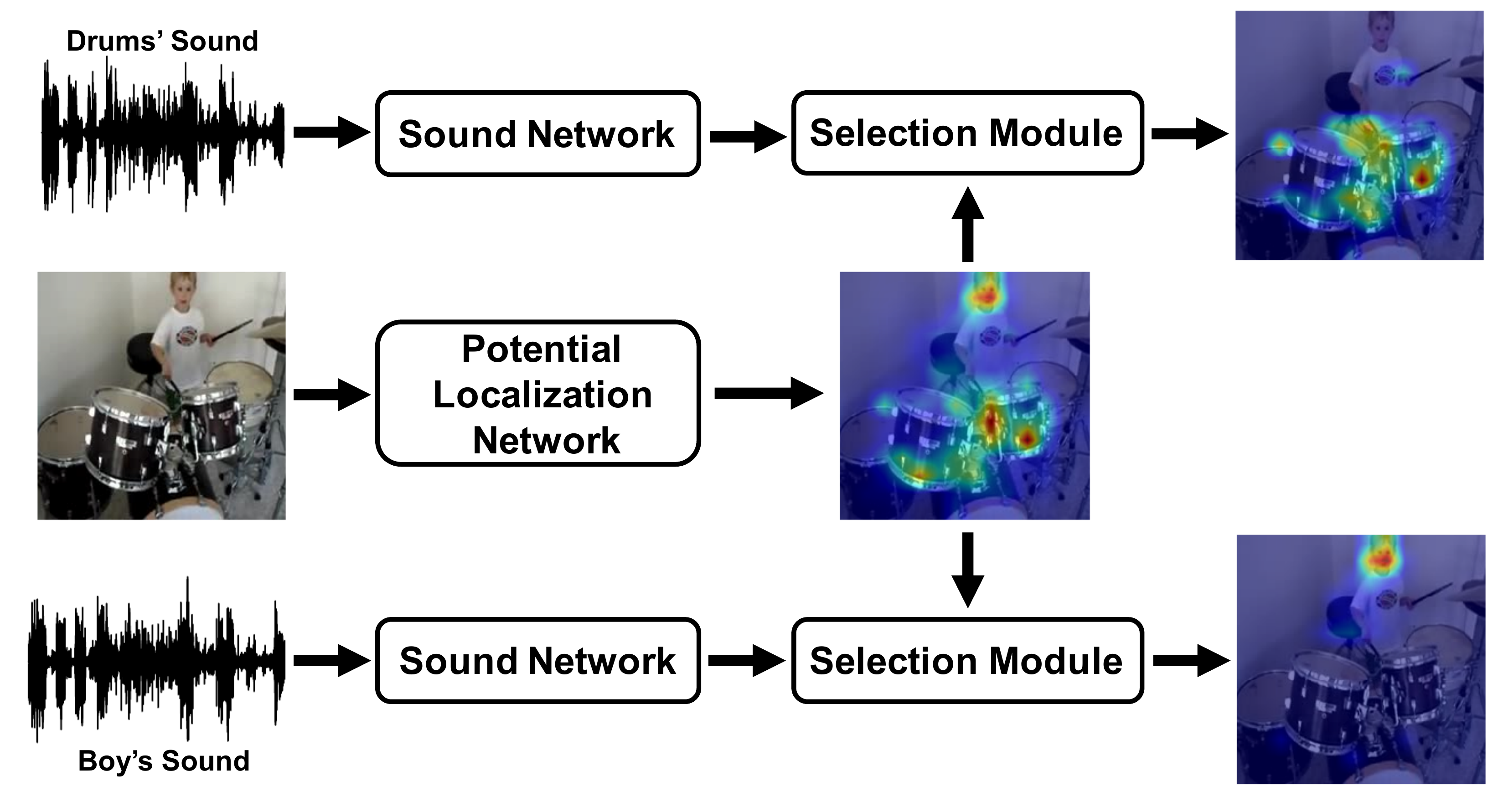}
    }
    \caption{
Overview of our system. The system decomposes sound source localization into two steps, and each step is carried out by a specific module: ``potential localization network" and ``selection module". More specifically, the potential localization network obtains a heat map of all objects that are capable of producing sound, and given sound features from a sound network, the selection module identifies objects that are actually producing sound. In the example above, as the boy and drums are capable of producing sound, the potential localization network obtains a heat map that responds to their corresponding regions. Then, when human voice is given as input audio, the pixels that correspond to humans are selected by the selection module. On the other hand, when the sound of drums is given, the pixels that correspond to drums are selected.
    }
    \label{fig:concept}
\end{figure}

%% file: sec_02_related_work.tex
\section{Related Work}
Noting that a novel solution may find application in several domains, below we provide a list of cross-modal perception tasks that are closely related to the problem of sound source localization. Each task is accompanied with a description of how they are related. 
It should be noted that several tasks are sometimes studied in a single paper. 
\\
\\
\noindent
\textbf{Audio-Visual Correspondence (AVC)} 
Audio and visual events tend to occur concurrently within our daily lives. 
Additionally, humans learn to localize a sound source without any supervision by experiencing these natural co-occurrences a sufficient number of times since they were born. 
Like humans, in order for a machine to learn from these natural co-occurrences, Arandjelovic \& Zisserman \cite{LLL} proposed an AVC task that predicts whether a pair of sound clip and an image correspond with one another.
Though the AVC task itself does not have a practical application, the concept of AVC facilitates solutions for other cross-modal perception tasks in an unsupervised fashion, such as audio-visual representation learning \cite{LLL, objects_that_sound, deep_multimodal_clustering}, audio-visual source separation \cite{co_segmentation, sound_of_pixels, multisensory}, and sound source localization \cite{objects_that_sound, learning_to_localize, sound_of_pixels, sound_of_motions}.
Although our work does not directly map to the AVC task, it indirectly leverages this idea in an unsupervised setting. 
\\
\\
\noindent
\textbf{Audio-Visual representation learning} 
This task aims to generate representations of image/sound; these representations are useful for image/sound classification, zero-shot detection, cross-modal retrieval and action recognition. 
Audio-visual representation learning models can be grouped within 3 categories: the use of image features to train a sound network in a supervised manner \cite{see_hear_read, soundnet, word_unit, spoken_language}, the use of sound features to train an image network in a supervised manner \cite{LLL, see_hear_read, ambient_sound}, and via unsupervised methods \cite{objects_that_sound, multisensory, event_localization, cooperative_learning, deep_multimodal_clustering} including networks to classify motion information.
Though the aim of representation learning is different from that of sound source localization, some existing works have reported that the internal representation i.e., image features and sound features of the models developed for sound source localization is useful for image/sound classification and action recognition tasks \cite{multisensory, cooperative_learning, deep_multimodal_clustering}.
Our efforts are different from this task, as we do not focus on internal representations.
\\
\\
\noindent
\textbf{Audio-visual source separation.} 
Audio-visual source separation is a task that utilizes visual information to guide sound source separation, in contrast to blind sound source separation that does not use visual information \cite{blind_old, blind1, blind2, blind3}.
There are various methods employed for audio-visual source separation: NMF \cite{motion_informed, two_multimodal, learning_to_separate}, subspace methods \cite{independent_components, sparsity}, the mix-and-separate method \cite{sound_of_pixels, multisensory, co_segmentation, co_separating, sound_of_motions}, and the use of facial information for speech separation \cite{cocktail, visual_speech_enhancement, conversation_enhancement, blind_audiovisual_source_separation}.
Many works have attempted simultaneous sound source separation and sound source localization \cite{sound_of_pixels, multisensory, co_segmentation, sound_of_motions}, as it is important to identify which objects are producing sound to perform sound source separation.
For example, Zhao et al. \cite{sound_of_pixels} proposed a method that assigns sounds to each image pixel and can localize a sound source by visualizing the volume of each image pixel.
However, these previous works applied their methods to datasets containing a limited number of categories, such as a music instrument or speech.
In contrast, our work does not focus on sound source separation and can be applied to a noisy dataset containing an unconstrained number of categories.
\\
\\
\noindent
\textbf{Sound source localization}
Sound source localization is a classic problem in both science and engineering.
In robotics and signal processing research, sound source localization mainly indicates a task that identifies the spatial locations of the sound sources using only aural information from several microphones \cite{robotics2, robotics, robotics3, robotics4}.
On the contrary, in computer vision research, sound source localization indicates a task seeking to determine image pixels that are associated with sound sources, usually employing both visual and aural information from a single microphone, i.e., monaural hearing.
There exist various studies concerning sound source localization in the computer vision domain, including works that are not based on neural network techniques: mutual information and CCA \cite{synchrony, pixels_that_sound}, subspace methods \cite{statistical_models}, and keypoints \cite{harmony_in_motion}.
After the great success of neural networks in the year 2012 \cite{alexnet}, a growing number of works have been conducted that leverage neural network techniques, including the following methods: those that leverage motion information \cite{event_localization, sound_of_motions}, CAM-based methods \cite{multisensory, cooperative_learning}, methods based on attention mechanism \cite{objects_that_sound, learning_to_localize}, and methods that can perform concurrent sound source localization and sound source separation \cite{sound_of_pixels, multisensory, co_segmentation, sound_of_motions}.
A work that is most similar to the one described in this study is that of Senocak et al. \cite{learning_to_localize}, which can be trained in an unsupervised manner and can be applied to a noisy, unlabelled dataset. However, the focus of their work is different from ours, because our study's aim is to investigate the contributions of each modality in sound source localization. Furthermore, our network differs from theirs in that we introduce potential sound source localization network and object selection module.

%% file: sec_03_proposed_framework.tex
\section{Proposed Framework}
\subsection{Dataset and Data Preprocessing} 
As a benchmark, we employ a dataset created by Senocak et al.\cite{learning_to_localize}; this dataset contains 5k random subset selections of Flickr-SoundNet \cite{soundnet}, a large unlabelled dataset containing more than 200M matched pairs of images and sounds. In addition, this dataset contains sound source annotations using bounding-boxes and labels indicating whether the sound is ambient or from a specific object; these labels were assigned by 3 annotators. 
To be accurate, annotators listened to the first 10 seconds of each audio clip and determined whether the clip contained either ambient or object sound. The annotators then performed bounding-box annotation of sound sources within the image. 
The image sizes within the dataset are fixed at 256 $\times$ 256 pixels, the sampling rate of the sound is fixed at 22.05 kHz, and the length of the sound clips is not fixed.

For image data, we performed a simple preprocessing and augmentation scheme: mean subtraction and random cropping (224 $\times$ 224). For audio data, we performed STFT with a window-length of 1022 and a hop-length of 511.

\subsection{Potential localization network}
The overall network architecture is given in Fig. \ref{fig:architecture}.
The potential localization network takes image vectors $I_{i, j} \in R^{H \times W}$ with height $H$ and width $W$ as inputs. Then, a VGG-11 network \cite{VGG} pretrained on ImageNet \cite{imagenet} and two $1 \times 1$ convolution layers followed by ReLU \cite{ReLU} were used to extract a positive vector $v_{k, i, j} \in R^{K \times \frac{H}{16} \times \frac{W}{16}}$ that represents image features with K channels. Finally, a potential localization map $P_{i, j} \in R^{\frac{H}{16} \times \frac{W}{16}}$ is computed by applying a $1 \times 1$ convolution layer followed by a softmax function, as shown below.
\begin{align}
    P_{i, j} = \frac{\exp[{\sum_k({W_k v_{k, i, j}}) + b}]}{\sum_{i, j} \exp[{\sum_k({W_k v_{k, i, j}}) + b}]}.
\end{align}
    Here $W_k, b$ are weight and bias term of the $1 \times 1$ convolution layer, respectively.

\subsection{Sound network}
The sound network takes the amplitudes and phases of a spectrogram as an input. This information is then fed into a VGG-like Network. Here, the sound length is arbitrary, caused by the application of a global average pooling (GAP) layer \cite{NIN} along the time axis after the VGG-like network has acted on the data.  Then, the positive vector $s_k \in R^K$ is obtained which represents the sound features after having been fed through two fully-connected (FC) layers followed by a ReLU.

\input{fig_architecture.tex}

\subsection{Selection module}
The selection module takes 3 inputs: image features $v_{k, i, j}$, sound features $s_k$, and the potential localization map $P_{i, j}$. An attention map $A_{i, j}$ is obtained by calculating the cosine similarity between image features $v_{k, i, j}$, and sound features $s_k$, as shown below:
\begin{align}
    A_{i, j} = \frac{\sum_k v_{k, i, j} s_k}{\sqrt{\sum_k s_k^2} \sqrt{\sum_k v_{k, i, j}^2}}.
\end{align}
    It should be noted that $A_{i, j}$ ranges between 0 and 1, as both  $v_{k, i, j}$ and $s_k$ are positive.
    Finally, the localization map $\alpha_{i, j}$ is calculated by L1 normalization along pixels after multiplying $A_{i, j}$ and $P_{i, j}$ together as shown by the following expression:
\begin{align}
    \alpha_{i, j} = \frac{A_{i, j} P_{i, j}}{\sum_{i, j} A_{i, j} P_{i, j}}.
\end{align}
\subsection{Training phase}
\label{sec:training}
 For training, we leverage concepts from an AVC task \cite{LLL} in which networks are trained to determine whether data pairs (image and sound) correspond to each other.
Our loss function which we call ``similarity loss''  is calculated as follows.
We first calculated $\hat{s_k} \in R^K$ by 
\begin{align}
    \hat{s_k} = {\rm ReLU}({\rm FC}({\rm ReLU}({\rm FC}(\sum_{i, j} v_{k, i, j} \alpha_{i, j})))).
\end{align}
  Intuitively, $\hat{s_k}$ indicates image features $v_{k, i, j}$ that were filtered by the localization map $\alpha_{i, j}$.

    Then as in a Siamese network \cite{siamese}, the loss function is defined differently for the positive pairs (pairs corresponding to each other) and the negative pairs (pairs not corresponding to each other).
    
	For both positive and negative pairs, the cosine similarity between $\hat{s_k}$ and $s_k$ is used.
    For the positive pairs, the loss function is given as follows:
\begin{align}
    L = 1 - \frac{\sum_k s_k \hat{s_k}}{\sqrt{\sum_k s_k^2} \sqrt{\sum_k \hat{s_k}^2}}.
\end{align}
    For the negative pairs, the loss function is given by the following expression:
\begin{align}
    L = \frac{\sum_k s_k \hat{s_k}}{\sqrt{\sum_k s_k^2} \sqrt{\sum_k \hat{s_k}^2}}.
\end{align}
    The loss function forces $\hat{s_k}$ to be similar to sound features $s_k$ for positive pairs and vice versa for negative pairs.
    Additional details such as hyperparameters are provided in the supplementary material. 

\subsection{Inference phase}
As we have shown, the potential localization network does not depend on aural information and works independently from other components. Therefore, a potential localization map $P_{i, j}$ can be obtained using only visual information.
On the other hand, aural information is needed to obtain a localization map $\alpha_{i, j}$, since the attention map $A_{i, j}$ necessary to derive $\alpha_{i, j}$ depends on aural information.

%% file: fig_architecture.tex
\begin{figure}[h]%[thpb]
    \centering
    \includegraphics[width=\linewidth]{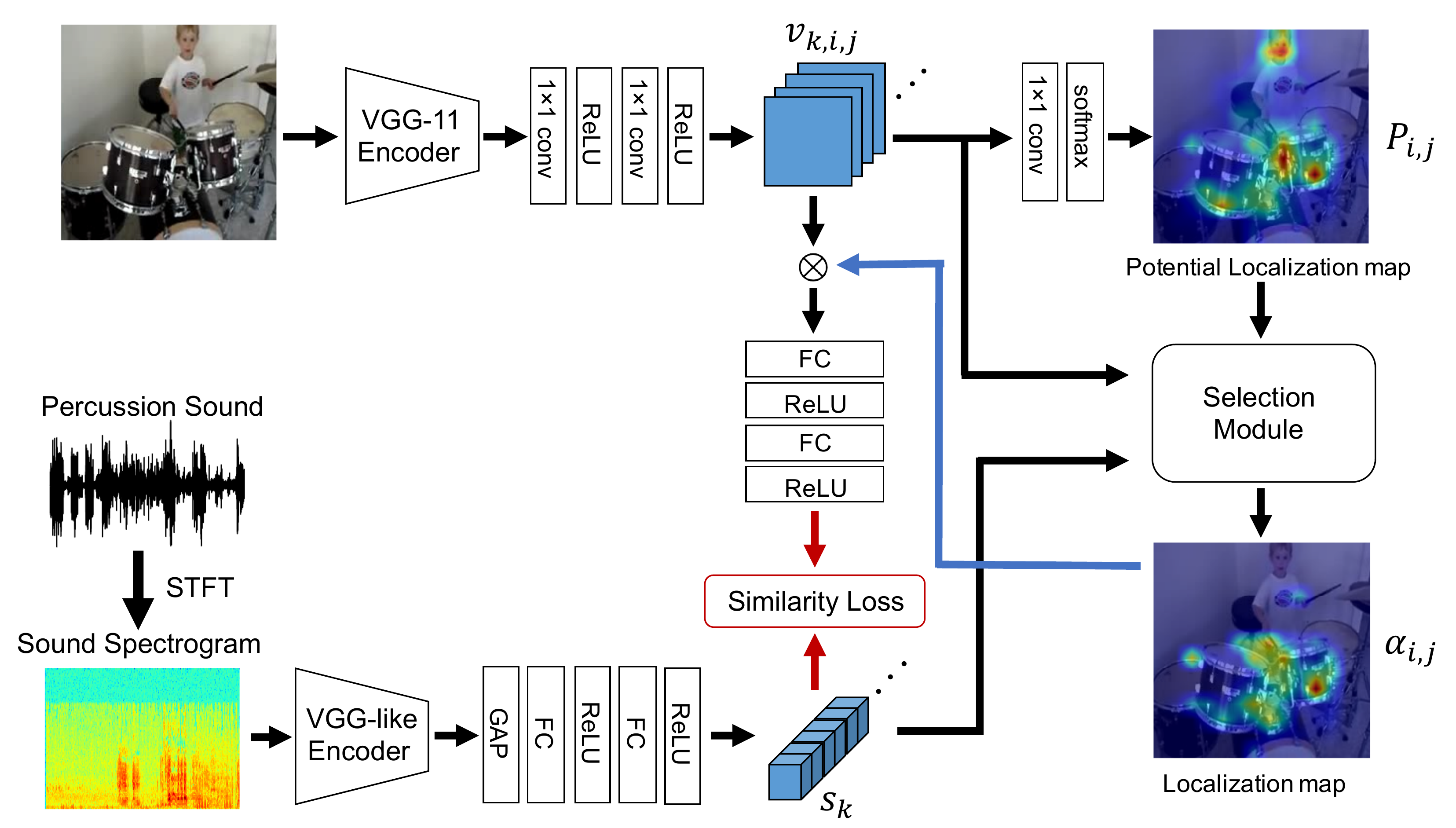}
    \caption{
    The detailed network architecture of our unsupervised model.  The potential localization map is obtained using image features extracted from the input image. Then from the potential localization map, the image features, and the sound features extracted from the input sound, the selection module produces a localization map. The similarity loss is calculated using sound features and image features of the sound-producing object obtained from the pixel-wise product between the localization map and the image features.
    }
    \label{fig:architecture}
\end{figure}

%% file: sec_04_experiments.tex
\section{Experiments}
We conducted 3 experiments. First, to verify the claim that potential localization maps respond to objects that potentially produce sound and not to objects that can not produce sound such as tables or chairs, we compared the results of the potential localization map to a saliency map. Then, to investigate the contributions of each modality, we trained our unsupervised method and compared the performance of (i) our potential localization map (ii) our localization map (iii) an existing unsupervised sound source localization method. Finally, we also conducted a similar evaluation in a supervised setting.   
\label{sec:experiments}

\subsection{Evaluation metrics}
To evaluate our system, we employ the consensus intersection over union (cIoU) \cite{learning_to_localize}, which is similar to intersection over union (IoU), but can be applied when ground truth takes continuous values.
The ground truth $G_{i, j}$ is given by
\begin{align}
G_{i, j} = \sum_{k = 1}^M \frac{{G_{k, i, j}}}{M}.
\end{align}
Here, $M$ indicates the number of annotators and $G_{k, i, j}$ indicates ground truth by k-th annotator, which takes 1 if the bounding box includes $(i, j)$-th pixel, and takes 0 otherwise. 
The definition of cIoU is as follows.
\begin{align}
{\rm cIoU} = \frac{\sum_{i, j} G_{i, j} B_{i, j}}{\sum_{i, j} G_{i, j} + \sum_{(i, j) \in \{(i, j) | G_{i, j} = 0\}} B_{i, j}},
\end{align}
where $B_{i, j}$ is obtained by binarizing the prediction $P_{i, j}$ using a threshold $\tau$.
\begin{align}
  B_{i, j} = \begin{cases}
    1 & (P_{i, j} > \tau) \\
    0 & (otherwise)
  \end{cases}.
\end{align}
Intuitively, the numerator of cIoU indicates the intersection of the ground truth and the prediction, and the denominator indicates the union. Therefore, a higher cIoU score indicates better performance. We also report AUC score, which indicates the area under the curve of cIoU plotted by varying the threshold from 0 to 1.
For evaluation, we use the same test set as Senocak et al. \cite{learning_to_localize}, which contains 250 pairs of image and sound.
\subsection{Configurations}
\noindent
\textbf{Saliency Map} 
We obtain the saliency map for the top-1 predicted class and the top-100 predicted classes of the pretrained VGG network using Grad-CAM \cite{grad_cam}.
We conducted this experiment to address the question: is a potential localization map just a saliency map, thus visualizing the important region for image classification?
This experiment allows us to differentiate whether potential localization maps respond only to objects that potentially produce sound or to any given object within the image, including those that can not produce sound. Additional details of this experiment are given in the supplementary material.
\\
\\
\noindent
\textbf{Unsupervised learning} 
We trained our unsupervised method with various numbers of training samples (1k, 2.5k, 10k, 144k) using pretrained weights for the VGG network.
As a means of comparison, both the potential localization map and the localization map were evaluated to investigate the degree to which aural information contributes to the overall performance in the benchmark dataset.
We also trained our unsupervised method without pretrained weights, in order to make a fair comparison with results reported in the existing study \cite{learning_to_localize}, which does not use pretrained weights.
In addition, based on this experiment, we can verify whether our model can be trained without pretrained weights.
\\
\\
\noindent
\textbf{Supervised learning}  
To also investigate the contribution of visual information in a supervised setting, we trained a U-Net \cite{UNet} with the ground truth $G_{i, j}$ as a target and use Dice-coefficient as a loss function.
Dice-coefficient is defined as
\begin{align}
{\rm DC} = \frac{\sum_{i, j} G_{i, j} U_{i, j}}{\sum_{i, j} G_{i, j} + \sum_{i, j} U_{i, j}},
\end{align}
where $U_{i, j}$ indicates the prediction of U-Net.
Additional details including hyperparameters are given in the supplementary material.

\input{tab_cIoU.tex}
\input{tab_sup.tex}
\input{tab_saliency.tex}
\input{fig_vs_cam.tex} 
\input{fig_potential_map.tex}
\input{fig_select.tex} 

\subsection{Results and Analysis}
\label{sec:analysis}
Table \ref{tab:cIoU} and Table \ref{tab:sup} show the results reported by Senocak et al. \cite{learning_to_localize} and the results of our potential localization map and localization map in an unsupervised setting and a supervised setting respectively. 
Table \ref{tab:saliency} shows results for the saliency maps. 
Insights obtained from these results are as follows.
\\
\\
\\
\\
\noindent
\textbf{Behavioral analysis of potential localization map} 
As you can see from Table \ref{tab:cIoU} and Table \ref{tab:saliency}, when more than 10k samples are used to conduct unsupervised training with pretrained weights, and when more than 144k samples are used to conduct training without pretrained weights, our potential localization map outperforms the baselines of saliency maps.
This implies our potential localization map technique, trained with a sufficient number of samples can successfully identify objects that potentially produce sound and not just visualize a saliency map.
As can be seen in Figure \ref{fig:vs_cam_re}, it can be confirmed that our potential localization map focuses on objects that potentially produce sound, while a saliency map does not.  
\\
\\
\noindent
\textbf{Assessment of individual modalities' importance } 
When the quantity of training data is within the range 10k to 144k, our localization maps achieve state-of-the-art performance (68.4\%, 56.8\% in cIoU).
More notably, our potential localization map performs comparably to the localization map which uses both visual and aural information (1.6\% gap in cIoU for 144k training samples with pretrained weights). The difference in performance is consistent in the supervised setting (1.1\% gap in cIoU). This suggests that although sound source localization is a cross-modal task, when evaluated using the current benchmark dataset, visual information is dominant and the combined advantage of having both visual and aural information is not fully assessed.
\input{tab_type.tex}
\\
\\
\noindent
\textbf{Analysis of the performance gap} 
In order to investigate the difference in behavior between the potential localization map and the localization map, we make qualitative observations as shown in Figure \ref{fig:potential_map} and Figure \ref{fig:select}.
As seen from Figure \ref{fig:potential_map}, it can be observed that in an unsupervised setting, when all objects capable of producing sound are actually producing sound, the potential localization map performs similarly to the localization map. We call this type of sample Type A. On the other hand, when only some of the objects capable of producing sound are actually producing sound, such as in Figures \ref{fig:select}, the localization map performs better than the potential localization map. We call this type of sample Type B. 

To quantitatively evaluate this difference in behavior between samples that are Type A and Type B, we manually picked out 30 samples for each type from Flickr-SoundNet \cite{soundnet} excluding the samples contained in the current benchmark dataset \cite{learning_to_localize}, and then annotated sound sources using bounding-boxes. Further details on the annotation process are given in the supplementary material. IoU and AUC scores for both types were calculated to check which type of samples leads to the difference in performance between potential localization map and localization map. The results are shown in Table \ref{tab:type}. In accordance with our qualitative observation, the results show that the performance gap is smaller in Type A and larger in Type B. \\ \indent
Based on the differences seen between samples of Type A and Type B, it can be implied that the lack of performance gap when evaluated with the current benchmark dataset is caused by the majority of samples within the dataset being Type A. This indicates that, inherently, the majority of sound sources can be inferred based on visual information only, thus confirming that the contribution of aural information is minimal in the current benchmark dataset.  This is problematic as a proper evaluation of whether a system is capable of leveraging aural information can not be conducted. It is important to realize that in the real world, there are many examples where objects that are capable of producing sound are silent, and thus the usage of sound would be inevitable. Hence, datasets used for evaluation of sound source localization should necessitate the usage of sound.

%% file: tab_cIoU.tex
\begin{table}[t]
    \caption{Evaluation of localization map and potential localization map in an unsupervised setting. cIoU score with threshold 0.5 and AUC score are calculated for different sized training samples. We also report the results of random prediction. ``de novo'' indicates training without pretrained weights, and localization map is abbreviated as loc. map. Additional comparison with other methods are listed in the supplementary material.}
    \centering
    \scalebox{1}{
\begin{tabular}{cccccccccccccccc}
\hline
\begin{tabular}[c]{@{}c@{}}\# of \\ training\\ samples\end{tabular} &  & \multicolumn{2}{c}{\begin{tabular}[c]{@{}c@{}}loc. map \\ by \cite{learning_to_localize}\\ (reported)\end{tabular}} &  & \multicolumn{2}{c}{\begin{tabular}[c]{@{}c@{}}loc. map\\ (pretrain.)\end{tabular}} &  & \multicolumn{2}{c}{\begin{tabular}[c]{@{}c@{}}potential \\ loc. map\\ (pretrain.)\end{tabular}} &  & \multicolumn{2}{c}{\begin{tabular}[c]{@{}c@{}}loc. map \\ (de novo)\end{tabular}} &  & \multicolumn{2}{c}{\begin{tabular}[c]{@{}c@{}}potential\\  loc. map \\ (de novo)\end{tabular}} \\ \cline{3-4} \cline{6-7} \cline{9-10} \cline{12-13} \cline{15-16} 
                                                                    &  & cIoU                                            & AUC                                             &  & cIoU                                     & AUC                                     &  & cIoU                                           & AUC                                            &  & cIoU                                    & AUC                                     &  & cIoU                                           & AUC                                           \\ \hline
1k                                                                  &  & ---                                             & ---                                             &  & 48.7                                     & 46.4                                    &  & 45.9                                           & 43.8                                           &  & 36.5                                    & 34.1                                    &  & 35.3                                           & 33.4                                          \\
2.5k                                                                &  & ---                                             & ---                                             &  & 50.3                                     & 47.7                                    &  & 48.1                                           & 46.0                                           &  & 40.7                                    & 37.3                                    &  & 38.5                                           & 36.8                                          \\
10k                                                                 &  & 43.6                                            & 44.9                                            &  & 56.8                                     & 50.7                                    &  & 53.9                                           & 48.6                                           &  & 48.4                                    & 45.3                                    &  & 47.4                                           & 44.6                                          \\
144k                                                                &  & 66.0                                            & 55.8                                            &  & 68.4                                     & 57.0                                    &  & 66.8                                           & 56.2                                           &  & 66.7                                    & 56.3                                    &  & 65.5                                           & 55.5                                          \\ \hline
\multirow{2}{*}{Random}                                             &  &                                                 &                                                 &  &                                          & \multicolumn{3}{c}{cIoU}                                                                    & \multicolumn{3}{c}{AUC}                                                                     &                                         &  &                                                &                                               \\ \cline{7-12}
                                                                    &  &                                                 &                                                 &  &                                          & \multicolumn{3}{c}{34.1}                                                                    & \multicolumn{3}{c}{32.3}                                                                    &                                         &  &                                                &                                               \\ \hline
\end{tabular}    }
    \label{tab:cIoU}
\end{table}

%% file: tab_sup.tex
\begin{table}[t]
    \caption{Evaluation of potential localization map in a supervised setting, i.e. prediction of U-Net which only uses visual information. We use 2.5k samples for training as in the supervised setting of Senocak et al. \cite{learning_to_localize}. }
    \centering
    \scalebox{1}{
\begin{tabular*}{7cm}{@{\extracolsep{\fill}}clclc}
\hline
Method                           &  & cIoU &  & AUC  \\ \hline
loc. map by \cite{learning_to_localize} (reported) &  & 80.4 &  & 60.3 \\
potential loc. map               &  & 79.3 &  & 60.9 \\ \hline
\end{tabular*}    }
    \label{tab:sup}
\end{table}

%% file: tab_saliency.tex
\begin{table}[t]
    \caption{Evaluation of saliency maps for the top 1 class and the top 100 classes.}
    \centering
    \scalebox{1}{
\begin{tabular*}{7cm}{@{\extracolsep{\fill}}ccc}
\hline
Method                                                                     & \multicolumn{1}{c}{cIoU}
	& \multicolumn{1}{c}{AUC} \\ \hline
\begin{tabular}[c]{@{}l@{}}saliency map for the top1 class\end{tabular}     & 45.7 & 41.1                 \\
\begin{tabular}[c]{@{}l@{}}saliency map for the top100 classes\end{tabular} & 52.7 & 46.3                \\ \hline
\end{tabular*}    }
    \label{tab:saliency}
\end{table}

%% file: fig_vs_cam.tex
\begin{figure}[t]
    \scalebox{0.80}{
        \begin{tabular}{p{7.3em} p{7.3em} p{7.3em} p{0.16em} p{7.3em} p{7.3em} p{7.3em}}

        %\toprule
        %-------
            %-------
        %-------
             \hfil Original Image \hfil 
            & \hfil Potential \hfil 
            & \hfil Grad-CAM for \hfil
            & \hfil \hfil
            & \hfil Original Image\hfil 
            & \hfil Potential \hfil
            & \hfil Grad-CAM for\hfil \\
            \hfil \hfil 
            & \hfil Loc. map \hfil 
            & \hfil 'castle' \hfil
            & \hfil \hfil 
            & \hfil \hfil
            & \hfil Loc. map \hfil 
            & \hfil 'barbell' \hfil \\
        %-------
            %-------
            % 1st
            \begin{minipage}{7.3em}
                \centering
                \scalebox{0.418}
                {\includegraphics{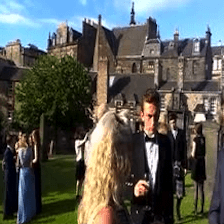}}
            \end{minipage} &
            % 2nd
            \begin{minipage}{7.3em}
                \centering
                \scalebox{0.418}
                {\includegraphics{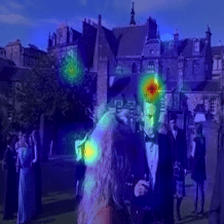}}
            \end{minipage} &
            % 3rd
            \begin{minipage}{7.3em}
                \centering
                \scalebox{0.418}
                {\includegraphics{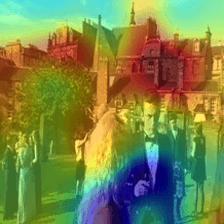}}
            \end{minipage} &
         	\begin{minipage}{0.16em}
         	\end{minipage} &
            % 4th
            \begin{minipage}{7.3em}
                \centering
                \scalebox{0.418}
                {\includegraphics{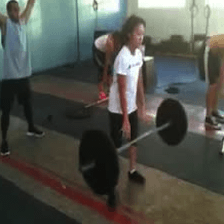}}
            \end{minipage} &
            
            \begin{minipage}{7.3em}
                \centering
                \scalebox{0.418}
                {\includegraphics{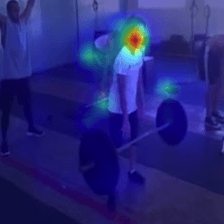}}
            \end{minipage} &
            % 4th
            \begin{minipage}{7.3em}
                \centering
                \scalebox{0.418}
                {\includegraphics{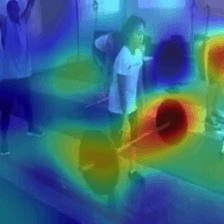}}
            \end{minipage} \vspace{1mm} \\
             \hfil Original Image \hfil 
            & \hfil Potential \hfil 
            & \hfil Grad-CAM for \hfil
            & \hfil \hfil 
            & \hfil Original Image \hfil 
            & \hfil Potential \hfil
            & \hfil Grad-CAM for \hfil \\
            \hfil \hfil 
            & \hfil Loc. map \hfil 
            & \hfil 'volcano' \hfil 
            & \hfil \hfil
            & \hfil \hfil
            & \hfil Loc. map\hfil 
            & \hfil 'china cabinet' \hfil \\
        %-------
            % 1st
            \begin{minipage}{7.3em}
                \centering
                \scalebox{0.418}
                {\includegraphics{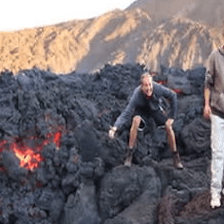}}
            \end{minipage} &
            % 2nd
            \begin{minipage}{7.3em}
                \centering
                \scalebox{0.418}
                {\includegraphics{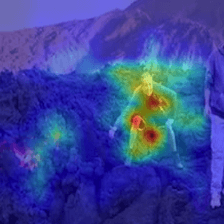}}
            \end{minipage} &
            % 3rd
            \begin{minipage}{7.3em}
                \centering
                \scalebox{0.418}
                {\includegraphics{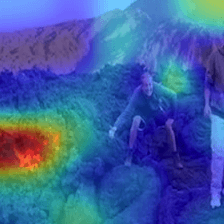}}
            \end{minipage} &
         	\begin{minipage}{0.16em}
         	\end{minipage} &
            % 4th
            \begin{minipage}{7.3em}
                \centering
                \scalebox{0.418}
                {\includegraphics{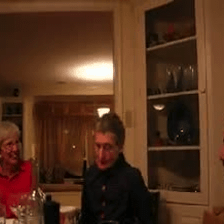}}
            \end{minipage} &
            
            \begin{minipage}{7.3em}
                \centering
                \scalebox{0.418}
                {\includegraphics{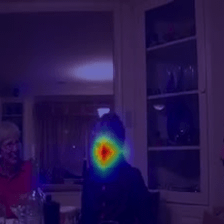}}
            \end{minipage} &
            % 4th
            \begin{minipage}{7.3em}
                \centering
                \scalebox{0.418}
                {\includegraphics{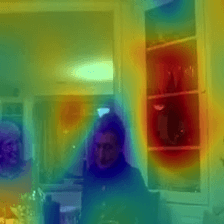}}
            \end{minipage} \\
        %\bottomrule
        %-------
        \end{tabular}     }
		\quad
    	\caption{Visualization results of potential localization map and saliency map (Grad-CAM) for the top 1 class. As the results show, the saliency maps react to objects that can not produce sound like ``china cabinet'', while our potential localization map does not. For the visualization, we use the model trained with 144k samples and pretrained weights.}
    \label{fig:vs_cam_re}
\end{figure}

%% file: fig_potential_map.tex
\begin{figure}[h]
    \scalebox{0.775}{
        \begin{tabular}{p{7.6em} p{7.6em} p{7.6em} p{0.16em} p{7.6em} p{7.6em} p{7.6em}}
        %\toprule
            %-------
        %-------
            \hfil Original Image\hfil 
            & \hfil Potential \hfil 
            & \hfil Loc. map\hfil
            & \hfil \hfil 
            & \hfil Original Image\hfil
            & \hfil Potential \hfil 
            & \hfil Loc. map\hfil \\
                    \hfil \hfil 
            & \hfil Loc. map \hfil 
            & \hfil \hfil
            & \hfil \hfil 
            & \hfil \hfil
            & \hfil Loc. map \hfil 
            & \hfil \hfil \\
        %-------
            % 1st
            \begin{minipage}{7.6em}
                \centering
                \scalebox{0.431}
                {\includegraphics{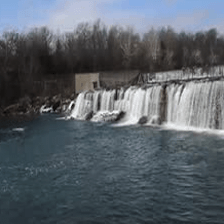}}
            \end{minipage} &
            % 2nd
            \begin{minipage}{7.6em}
                \centering
                \scalebox{0.431}
                {\includegraphics{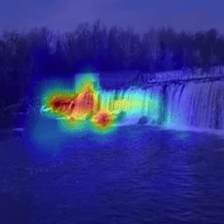}}
            \end{minipage} &
            % 3rd
            \begin{minipage}{7.6em}
                \centering
                \scalebox{0.431}
                {\includegraphics{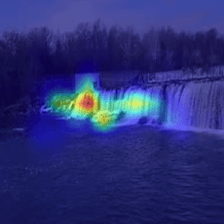}}
            \end{minipage} &
            \begin{minipage}{0.16em}
            \end{minipage} &
            % 4th
            \begin{minipage}{7.6em}
                \centering
                \scalebox{0.431}
                {\includegraphics{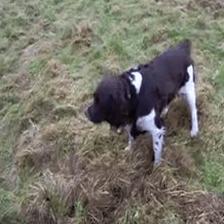}}
            \end{minipage} &
            
            \begin{minipage}{7.6em}
                \centering
                \scalebox{0.431}
                {\includegraphics{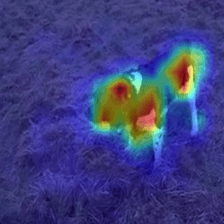}}
            \end{minipage} &
            % 4th
            \begin{minipage}{7.6em}
                \centering
                \scalebox{0.431}
                {\includegraphics{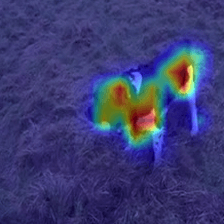}}
            \end{minipage} \vspace{0.001mm} \\
        %-------
        %-------
            %-------
            % 1st
            \begin{minipage}{7.6em}
                \centering
                \scalebox{0.431}
                {\includegraphics{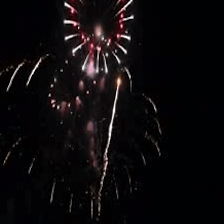}}
            \end{minipage} &
            % 2nd
            \begin{minipage}{7.6em}
                \centering
                \scalebox{0.431}
                {\includegraphics{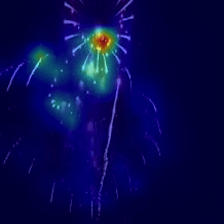}}
            \end{minipage} &
            % 3rd
            \begin{minipage}{7.6em}
                \centering
                \scalebox{0.431}
                {\includegraphics{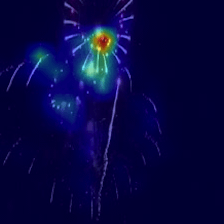}}
            \end{minipage} &
            \begin{minipage}{0.16em}
            \end{minipage} &
            % 4th
            \begin{minipage}{7.6em}
                \centering
                \scalebox{0.431}
                {\includegraphics{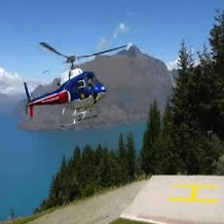}}
            \end{minipage} &
            
            \begin{minipage}{7.6em}
                \centering
                \scalebox{0.431}
                {\includegraphics{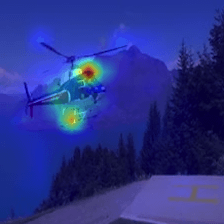}}
            \end{minipage} &
            % 4th
            \begin{minipage}{7.6em}
                \centering
                \scalebox{0.431}
                {\includegraphics{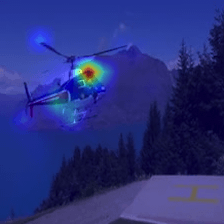}}
            \end{minipage} \vspace{0.001mm} \\
        %-------
            % 1st
            \begin{minipage}{7.6em}
                \centering
                \scalebox{0.431}
                {\includegraphics{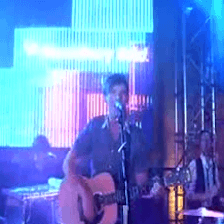}}
            \end{minipage} &
            % 2nd
            \begin{minipage}{7.6em}
                \centering
                \scalebox{0.431}
                {\includegraphics{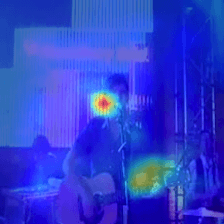}}
            \end{minipage} &
            % 3rd
            \begin{minipage}{7.6em}
                \centering
                \scalebox{0.431}
                {\includegraphics{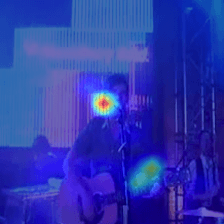}}
            \end{minipage} &
            \begin{minipage}{0.16em}
            \end{minipage} &
            % 4th
            \begin{minipage}{7.6em}
                \centering
                \scalebox{0.431}
                {\includegraphics{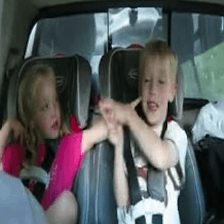}}
            \end{minipage} &
            
            \begin{minipage}{7.6em}
                \centering
                \scalebox{0.431}
                {\includegraphics{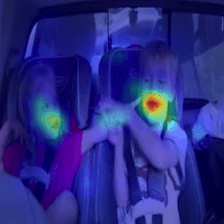}}
            \end{minipage} &
            % 4th
            \begin{minipage}{7.6em}
                \centering
                \scalebox{0.431}
                {\includegraphics{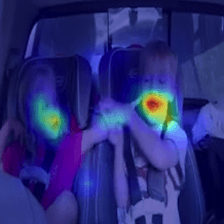}}
            \end{minipage} \vspace{0.001mm} \\
            %-------
        %-------
        %-------
            % 1st
            \begin{minipage}{7.6em}
                \centering
                \scalebox{0.431}
                {\includegraphics{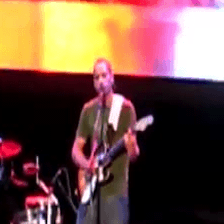}}
            \end{minipage} &
            % 2nd
            \begin{minipage}{7.6em}
                \centering
                \scalebox{0.431}
                {\includegraphics{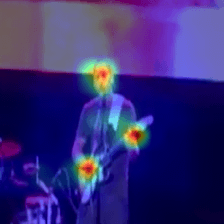}}
            \end{minipage} &
            % 3rd
            \begin{minipage}{7.6em}
                \centering
                \scalebox{0.431}
                {\includegraphics{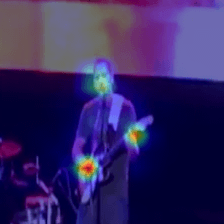}}
            \end{minipage} &
            \begin{minipage}{0.16em}
            \end{minipage} &
            % 4th
            \begin{minipage}{7.6em}
                \centering
                \scalebox{0.431}
                {\includegraphics{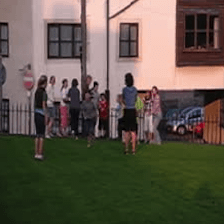}}
            \end{minipage} &
            
            \begin{minipage}{7.6em}
                \centering
                \scalebox{0.431}
                {\includegraphics{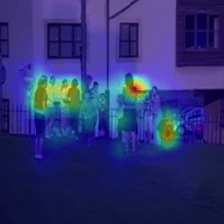}}
            \end{minipage} &
            % 4th
            \begin{minipage}{7.6em}
                \centering
                \scalebox{0.431}
                {\includegraphics{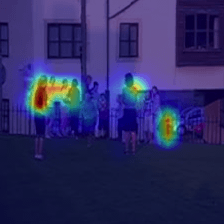}}
            \end{minipage} \\
        
        %\bottomrule
        %-------
        \end{tabular}     }
    	\caption{Visualization results of potential localization maps and localization maps of various objects when all possible objects are producing sound (Type A). Samples in the two top rows only have one object capable of producing sound. Samples in the bottom two rows have multiple objects but the sound of both objects are given. In these cases the results show that the potential localization map and the localization map are very similar. For the visualization, we use the model trained with 144k samples and pretrained weights.}
    \label{fig:potential_map}
\end{figure}

%% file: fig_select.tex
\begin{figure*}[h]
    \scalebox{0.56}{
        \begin{tabular}{p{7.8em} p{7.8em} p{7.8em} p{7.8em} p{0.18em} p{7.8em} p{7.8em} p{7.8em} p{7.8em}}
        %\toprule
        %-------
        	\hfil Original\hfil 
            & \hfil Potential \hfil 
            & \hfil Loc. map\hfil 
            & \hfil Loc. map\hfil
            & \hfil \hfil
            & \hfil Original\hfil 
            & \hfil Potential \hfil 
            & \hfil Loc. map\hfil 
            & \hfil Loc. map\hfil \\
            
             \hfil Image \hfil
            & \hfil Loc. map \hfil 
            & \hfil when given\hfil 
            & \hfil when given\hfil
            & \hfil \hfil
            & \hfil Image \hfil
            & \hfil Loc. map \hfil 
            & \hfil when given\hfil 
            & \hfil when given\hfil \\
            
         	 \hfil  \hfil 
         	& \hfil  \hfil 
            & \hfil human sound\hfil 
            & \hfil instrument sound\hfil
            & \hfil \hfil
            & \hfil  \hfil
            & \hfil \hfil 
            & \hfil human sound\hfil 
            & \hfil machine sound\hfil  \\
            \begin{minipage}{7.8em}
                \centering
                \scalebox{0.445}
                {\includegraphics{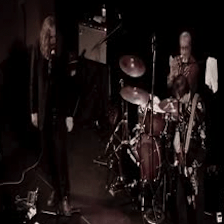}}
            \end{minipage} &        
            % 1st
            \begin{minipage}{7.8em}
                \centering
                \scalebox{0.445}
                {\includegraphics{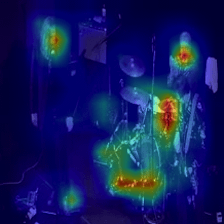}}
            \end{minipage} &
            % 2nd
            \begin{minipage}{7.8em}
                \centering
                \scalebox{0.445}
                {\includegraphics{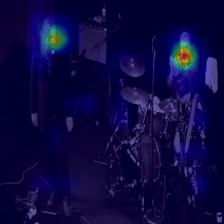}}
            \end{minipage} &
            % 3rd
            \begin{minipage}{7.8em}
                \centering
                \scalebox{0.445}
                {\includegraphics{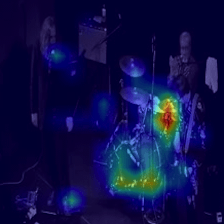}}
            \end{minipage} &
            \begin{minipage}{0.18em}
            \end{minipage} &
            \begin{minipage}{7.8em}
                \centering
                \scalebox{0.445}
                {\includegraphics{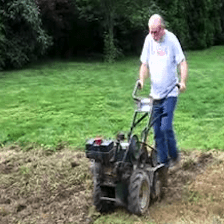}}
            \end{minipage} &
            % 4th
            \begin{minipage}{7.8em}
                \centering
                \scalebox{0.445}
                {\includegraphics{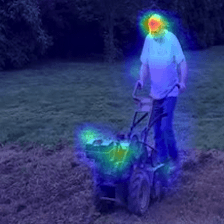}}
            \end{minipage} &
            
            \begin{minipage}{7.8em}
                \centering
                \scalebox{0.445}
                {\includegraphics{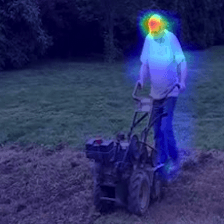}}
            \end{minipage} &
            % 4th
            \begin{minipage}{7.8em}
                \centering
                \scalebox{0.445}
                {\includegraphics{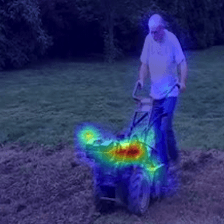}}
            \end{minipage} \vspace{0.001mm} \\
            %-------
        %-------
        %-------
            \begin{minipage}{7.8em}
                \centering
                \scalebox{0.445}
                {\includegraphics{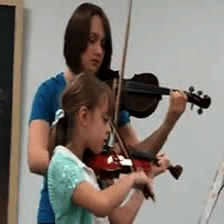}}
            \end{minipage} &
            % 1st
            \begin{minipage}{7.8em}
                \centering
                \scalebox{0.445}
                {\includegraphics{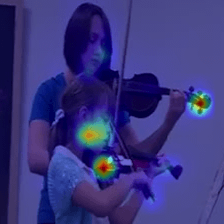}}
            \end{minipage} &
            % 2nd
            \begin{minipage}{7.8em}
                \centering
                \scalebox{0.445}
                {\includegraphics{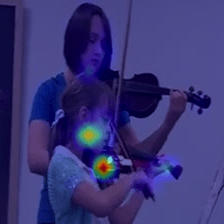}}
            \end{minipage} &
            % 3rd
            \begin{minipage}{7.8em}
                \centering
                \scalebox{0.445}
                {\includegraphics{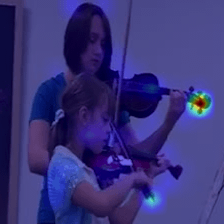}}
            \end{minipage} &
            \begin{minipage}{0.18em}
            \end{minipage} &
            \begin{minipage}{7.8em}
                \centering
                \scalebox{0.445}
                {\includegraphics{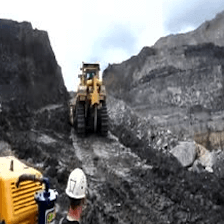}}
            \end{minipage} &
            % 4th
            \begin{minipage}{7.8em}
                \centering
                \scalebox{0.445}
                {\includegraphics{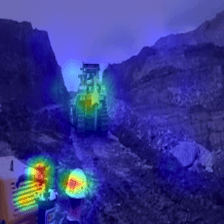}}
            \end{minipage} &
            
            \begin{minipage}{7.8em}
                \centering
                \scalebox{0.445}
                {\includegraphics{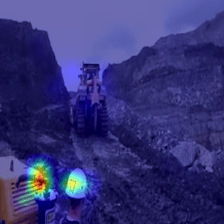}}
            \end{minipage} &
            % 4th
            \begin{minipage}{7.8em}
                \centering
                \scalebox{0.445}
                {\includegraphics{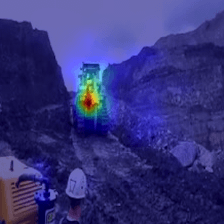}}
            \end{minipage} \vspace{0.001mm} \\
        %-------
        %-------
            %-------
            \begin{minipage}{7.8em}
                \centering
                \scalebox{0.445}
                {\includegraphics{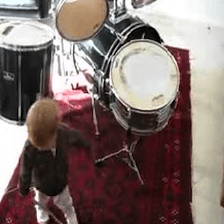}}
            \end{minipage} &
            % 1st
            \begin{minipage}{7.8em}
                \centering
                \scalebox{0.445}
                {\includegraphics{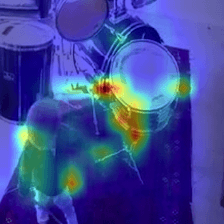}}
            \end{minipage} &
            % 2nd
            \begin{minipage}{7.8em}
                \centering
                \scalebox{0.445}
                {\includegraphics{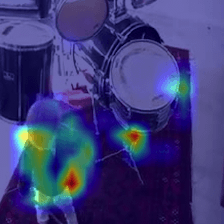}}
            \end{minipage} &
            % 3rd
            \begin{minipage}{7.8em}
                \centering
                \scalebox{0.445}
                {\includegraphics{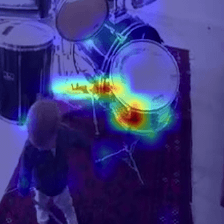}}
            \end{minipage} &
            \begin{minipage}{0.18em}
            \end{minipage} &
            \begin{minipage}{7.8em}
                \centering
                \scalebox{0.445}
                {\includegraphics{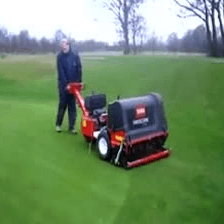}}
            \end{minipage} &
            % 4th
            \begin{minipage}{7.8em}
                \centering
                \scalebox{0.445}
                {\includegraphics{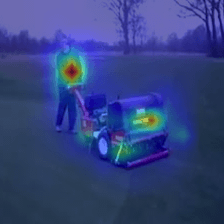}}
            \end{minipage} &
            
            \begin{minipage}{7.8em}
                \centering
                \scalebox{0.445}
                {\includegraphics{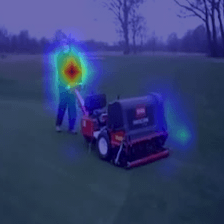}}
            \end{minipage} &
            % 4th
            \begin{minipage}{7.8em}
                \centering
                \scalebox{0.445}
                {\includegraphics{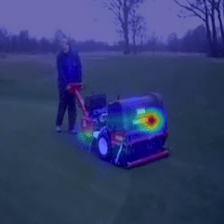}}
            \end{minipage} \vspace{0.001mm} \\
            %-------
        
        %\bottomrule
        %-------
        \end{tabular}     }
    	\caption{Visualization results of potential localization maps and localization maps of samples with several objects capable of producing sound in the image, but of which only one object is actually producing sound (Type B). The localization map successfully responds to an object that is actually producing sound, while the potential localization map does not. For the visualization, we use the model trained with 144k samples and pretrained weights.}
    \label{fig:select}
\end{figure*}

%% file: tab_type.tex
\begin{table}[t]
    \caption{IoU score with threshold 0.5 and AUC score for Type A and Type B samples. We use the model trained with 144k samples and pretrained weights.}
    \centering
    \scalebox{1}{
\begin{tabular*}{10cm}{@{\extracolsep{\fill}}ccllcllcllcl}
\hline
       & \multicolumn{2}{c}{loc. map} &  & \multicolumn{2}{c}{\begin{tabular}[c]{@{}c@{}}potential \\ loc. map\end{tabular}} &  & \multicolumn{2}{c}{\begin{tabular}[c]{@{}c@{}}perform. \\ gap\end{tabular}} &  & \multicolumn{2}{c}{Random} \\ \cline{2-3} \cline{5-6} \cline{8-9} \cline{11-12} 
       & IoU           & AUC          &  & IoU                                     & AUC                                     &  & IoU                                  & AUC                                 &  & IoU          & AUC         \\ \hline
Type A & 58.5          & 49.6         &  & 55.7                                    & 48.3                                    &  & 2.8                                  & 1.3                                 &  & 29.9         & 28.4        \\
Type B & 44.5          & 38.0         &  & 31.9                                    & 27.2                                    &  & 12.6                                 & 10.8                                &  & 17.1         & 16.6        \\ \hline
\end{tabular*}    }
    \label{tab:type}
\end{table}

%% file: sec_05_improved_eval.tex
\input{tab_concat.tex}

\section{Alternative Evaluation Protocol}
From our observations drawn in Sec. \ref{sec:analysis}, it can be deduced that a dataset which consists of abundant samples of Type B is required to properly assess a system's capability of leveraging aural information. However, as the retrieval of such samples in sufficient amounts is implausible, we introduce a {\it concat-and-localize} method, which is an alternative evaluation protocol that utilizes artificially created samples which structurally have the same properties as Type B.

Inspired by the mix-and-separate method used in audio-visual source separation \cite{sound_of_pixels, multisensory, co_segmentation, co_separating, sound_of_motions}, the process of artificially creating a sample in our {\it concat-and-localize} method is as follows.  First, two pairs of image and sound are sampled from the test set of the current dataset \cite{learning_to_localize}.  The two images are concatenated side-by-side and then rescaled to the size of the original image. Only one of the two sounds is given as input, and the new ground truth is defined as the rescaled original ground truth of the sample corresponding to the input sound as shown in Fig. \ref{fig:concat}. This process is conducted for both sounds, i.e. two synthetic samples are obtained per process. By design, the newly created sample always consists of multiple objects that are capable of producing sound as each original image contains such objects. In addition, as only one of the sounds is given, we can reproduce a situation where only some of the objects are producing sound, meeting the requirements for the sample to be Type B. Hence, in order to determine whether the sound originates from the objects on the left side of the concatenated image or the ones on the right side, aural information must be leveraged. \\ \indent
We created a new dataset of 1000 samples using the above process and evaluated our system using this dataset. The same metrics (cIoU and AUC) as Sec. \ref{sec:experiments} were utilized. In Table \ref{tab:concat_cIoU}, we show the results of our potential localization map and localization map following the same configuration as Sec. \ref{sec:experiments}. As expected, the results show that the performance gap is significantly greater than the currently adopted evaluation method, especially when training with a large number of samples (11.4\% gap in cIoU with 144k training samples). This can be attributed to the design of our evaluation protocol, where the problem cannot be solved using only visual information. In this respect, our {\it concat-and-localize} method is more suitable for evaluating cross-modal perception task than simply using the current benchmark dataset.
\input{fig_concat.tex} 

%% file: tab_concat.tex
\begin{table}[t]
    \caption{Evaluation of localization map and potential localization map in an unsupervised setting following our evaluation protocol. cIoU score with threshold 0.5 and AUC score are calculated for different sized training samples.}
    \centering
    \scalebox{1}{
\begin{tabular}{cccccccccccccccccc}
\cline{1-3} \cline{5-6} \cline{8-9} \cline{11-12} \cline{14-15} \cline{17-18}
\begin{tabular}[c]{@{}c@{}}\# of \\ training\\ samples\end{tabular} & \multicolumn{2}{c}{\begin{tabular}[c]{@{}c@{}}loc. map\\ (pretrain.)\end{tabular}} &  & \multicolumn{2}{c}{\begin{tabular}[c]{@{}c@{}}potential \\ loc. map\\ (pretrain.)\end{tabular}} &  & \multicolumn{2}{c}{\begin{tabular}[c]{@{}c@{}}perform. \\ gap\\ (pretrain.)\end{tabular}} &  & \multicolumn{2}{c}{\begin{tabular}[c]{@{}c@{}}loc. map\\ (de novo)\end{tabular}} &  & \multicolumn{2}{c}{\begin{tabular}[c]{@{}c@{}}potential \\ loc. map \\ (de novo)\end{tabular}} &  & \multicolumn{2}{c}{\begin{tabular}[c]{@{}c@{}}perform.\\  gap\\ (de novo)\end{tabular}} \\ \cline{2-3} \cline{5-6} \cline{8-9} \cline{11-12} \cline{14-15} \cline{17-18} 
                                                                    & cIoU                                     & AUC                                     &  & cIoU                                           & AUC                                            &  & cIoU                                         & AUC                                        &  & cIoU                                    & AUC                                    &  & cIoU                                           & AUC                                           &  & cIoU                                        & AUC                                       \\ \hline
1k                                                                  & 24.8                                     & 23.5                                    &  & 23.4                                           & 21.8                                           &  & 1.4                                          & 1.7                                        &  & 21.8                                    & 20.4                                   &  & 21.1                                           & 20.2                                          &  & 0.7                                         & 0.2                                       \\
2.5k                                                                & 27.3                                     & 24.4                                    &  & 23.5                                           & 21.3                                           &  & 3.8                                          & 3.1                                        &  & 23.2                                    & 21.5                                   &  & 22.9                                           & 20.8                                          &  & 0.3                                         & 0.7                                       \\
10k                                                                 & 35.4                                     & 29.5                                    &  & 28.2                                           & 25.1                                           &  & 7.2                                          & 4.4                                        &  & 30.1                                    & 26.6                                   &  & 24.0                                           & 22.1                                          &  & 6.1                                         & 4.5                                       \\
144k                                                                & 41.6                                     & 35.6                                    &  & 30.2                                           & 26.4                                           &  & 11.4                                         & 9.2                                        &  & 37.4                                    & 31.6                                   &  & 28.9                                           & 25.4                                          &  & 8.5                                         & 6.2                                       \\ \hline
\multirow{2}{*}{Random}                                             &                                          &                                         &  &                                                &                                                &  & \multicolumn{2}{c}{cIoU}                                                                  &  & \multicolumn{2}{c}{AUC}                                                          &  &                                                &                                               &  &                                             &                                           \\ \cline{8-12}
                                                                    &                                          &                                         &  &                                                &                                                &  & \multicolumn{2}{c}{20.0}                                                                  &  & \multicolumn{2}{c}{18.5}                                                         &  &                                                &                                               &  &                                             &                                           \\ \hline
\end{tabular}    }
    \label{tab:concat_cIoU}
\end{table}

%% file: fig_concat.tex
\begin{figure*}[t]
    \scalebox{0.765}{
        \begin{tabular}{p{7.8em} p{7.8em} p{7.8em} p{7.8em} p{7.8em} p{7.8em}}
        %\toprule
        %-------
             \hfil Concatenated  \hfil 
            & \hfil Potential \hfil 
            & \hfil Ground truth \hfil
            & \hfil Loc. map \hfil 
            & \hfil Ground truth \hfil 
            & \hfil Loc. map \hfil \\
            
             \hfil  Image \hfil 
            & \hfil Loc. map \hfil 
            & \hfil when given \hfil
            & \hfil when given \hfil 
            & \hfil when given \hfil 
            & \hfil when given \hfil \\

             \hfil  \hfil 
            & \hfil \hfil 
            & \hfil left sound \hfil
            &     \hfil left sound \hfil 
            & \hfil right sound \hfil 
            & \hfil right sound \hfil  \\
        %-------
        
            % 1st
            \begin{minipage}{7.8em}
                \centering
                \scalebox{0.435}
                {\includegraphics{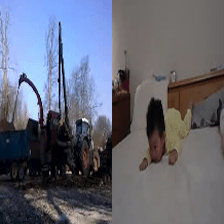}}
            \end{minipage} &
            % 2nd
            \begin{minipage}{7.8em}
                \centering
                \scalebox{0.435}
                {\includegraphics{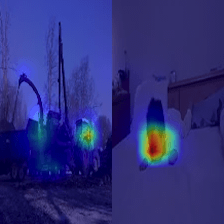}}
            \end{minipage} &
            % 3rd
            \begin{minipage}{7.8em}
                \centering
                \scalebox{0.435}
                {\includegraphics{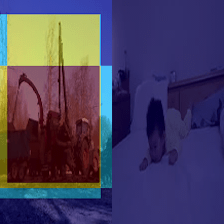}}
            \end{minipage} &
            % 4th
            \begin{minipage}{7.8em}
                \centering
                \scalebox{0.435}
                {\includegraphics{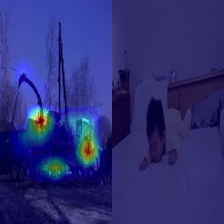}}
            \end{minipage} &
            
            \begin{minipage}{7.8em}
                \centering
                \scalebox{0.435}
                {\includegraphics{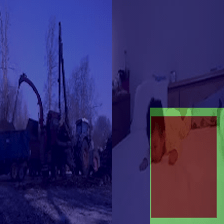}}
            \end{minipage} &
            % 4th
            \begin{minipage}{7.8em}
                \centering
                \scalebox{0.435}
                {\includegraphics{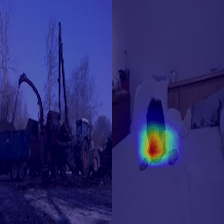}}
            \end{minipage} \vspace{0.001mm} \\
            %-------
        %-------
        %-------
            % 1st
            \begin{minipage}{7.8em}
                \centering
                \scalebox{0.435}
                {\includegraphics{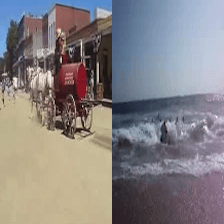}}
            \end{minipage} &
            % 2nd
            \begin{minipage}{7.8em}
                \centering
                \scalebox{0.435}
                {\includegraphics{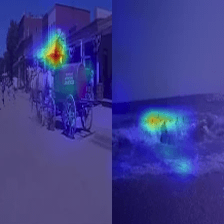}}
            \end{minipage} &
            % 3rd
            \begin{minipage}{7.8em}
                \centering
                \scalebox{0.435}
                {\includegraphics{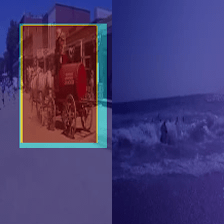}}
            \end{minipage} &
            % 4th
            \begin{minipage}{7.8em}
                \centering
                \scalebox{0.435}
                {\includegraphics{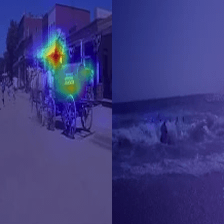}}
            \end{minipage} &
            
            \begin{minipage}{7.8em}
                \centering
                \scalebox{0.435}
                {\includegraphics{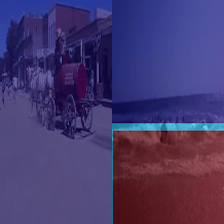}}
            \end{minipage} &
            % 4th
            \begin{minipage}{7.8em}
                \centering
                \scalebox{0.435}
                {\includegraphics{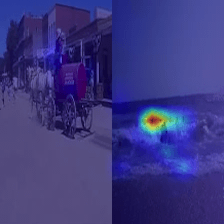}}
            \end{minipage} \vspace{0.001mm} \\
        %-------
        %-------
            %-------
            % 1st
            \begin{minipage}{7.8em}
                \centering
                \scalebox{0.435}
                {\includegraphics{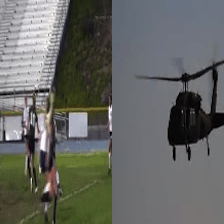}}
            \end{minipage} &
            % 2nd
            \begin{minipage}{7.8em}
                \centering
                \scalebox{0.435}
                {\includegraphics{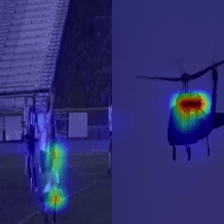}}
            \end{minipage} &
            % 3rd
            \begin{minipage}{7.8em}
                \centering
                \scalebox{0.435}
                {\includegraphics{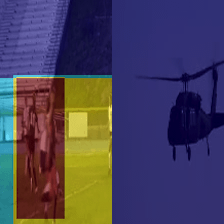}}
            \end{minipage} &
            % 4th
            \begin{minipage}{7.8em}
                \centering
                \scalebox{0.435}
                {\includegraphics{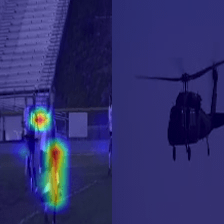}}
            \end{minipage} &
            
            \begin{minipage}{7.8em}
                \centering
                \scalebox{0.435}
                {\includegraphics{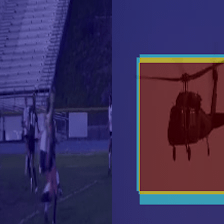}}
            \end{minipage} &
            % 4th
            \begin{minipage}{7.8em}
                \centering
                \scalebox{0.435}
                {\includegraphics{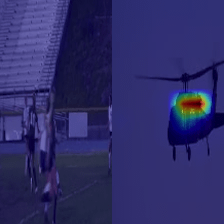}}
            \end{minipage} \\
            %-------
        
        %\bottomrule
        %-------
        \end{tabular}     }
    	\caption{Visualization results of potential localization maps and localization maps for artificially created samples. When given the concatenated image, the potential localization map shows response to all objects capable of producing sound in both images. Then when given only one of the sounds, the localization map correctly responds to the corresponding object.  For the visualization, we use the model trained with 144k samples and pretrained weights.}
    \label{fig:concat}
\end{figure*}

%% file: sec_06_conclusion.tex
\section{Conclusion}
In this paper, we develop an unsupervised architecture that solves sound source localization by decomposing this task into two steps: potential sound source localization and object selection. We provide sufficient evidence to show that our system can localize possible sound sources from only an image, and achieve state-of-the-art in sound source localization when leveraging both image and sound inputs. More importantly, our system, even when using only visual information, performs comparably to our state-of-the-art output which leverages both image and sound, suggesting sound sources can be localized without the usage of sound in the current dataset. From this observation and both qualitative and quantitative analysis, we pointed out the problem of the currently adopted evaluation method, and introduced the {\it concat-and-localize} method as an alternative. Our evaluation protocol is more suitable for evaluating the cross-modal aspect of sound source localization as the design enforces the usage of both modalities.

For future works, we believe it to be valuable to investigate the importance and contribution of different modalities in other cross-modal perception tasks too, as it may lead to a deeper understanding of the tasks, a dataset with better quality, and a better performance of future models in the real world.

%% file: supplement.tex
\label{sec:experiments}

\section{Unsupervised learning}
\noindent
\textbf{Training procedure} 
For the setting without pretrained weights for the image network, we use a fixed learning rate = 0.0001, with batchsize = 8 and epochs = 50.
When we use pretrained weights for the image network, we use batchsize = 8, epochs = 25 and learning rate = 0.00001 for the image network, and learning rate = 0.0001 for the others.
For all experiments, we use an Adam \cite{adam} optimizer.
\\
\\
\noindent
\textbf{Architecture details} 
For the image network, we use VGG-11, implemented in pytorch \cite{pytorch}. We show the architecture of the image network in Table \ref{tab:image_network}.
For the sound network, we use a VGG-like architecture. Table \ref{tab:sound_network} shows the architecture of the sound network. 

For more details of our unsupervised model, refer to our code included in the supplementary materials.

\section{Supervised learning}
For the supervised training, we train an U-Net \cite{UNet} with a fixed learning rate = 0.0001, batchsize = 8, epochs = 100, and the Adam optimizer.  The encoder of the U-Net is ResNet-34 \cite{resnet}. For More details of the architecture, refer to our code.
\input{tab_image_network}
\input{tab_sound_network}
\section{Saliency map}
We use the output of Grad-CAM \cite{grad_cam}, which can be regarded as a class-specific saliency map. To obtain the saliency map for top-N classes, we use the max value for each pixel along N saliency maps, as follows.
\begin{align}
S^N_{i, j} = \max_k S_{k, i, j},
\end{align}
where $S^N_{i, j}$ is the saliency map for top-N classes and $S_{k, i, j}$ is the saliency map for the k-th predicted class.

We test the performances of the saliency maps for top-N classes, for N = 1, 10, 20, 50, 100, 200, 500, 1000 and find N = 100 works best; thus we report the result of N = 100 in the original paper. Table \ref{tab:N_saliency} shows the performances of saliency maps using different N.
\input{tab_N_saliency}
%\input{tab_N_saliency}
%\section{Evaluation}
%To decide the threshold $\tau$ needed to calculate cIoU, we follow the procedure of Senocak et al. \cite{learning_to_localize}. To be more specific, We sort all the per-pixel-values of the predicted map $P_{i, j}$ and set the binarized predicted map $B_{i, j} = 1$ for the top 50\% of pixel values and set $B_{i, j} = 0$ otherwise. In other words, the threshold $\tau$ is the median of $P_{i, j}$.
\input{tab_comparison}
\section{Additional comparison with other methods}
Though our main focus is to analyze the contribution of image/sound modalities, we made further comparisons with other various methods\cite{deep_multimodal_clustering, curriculum, two_stage}. The results in \ref{tab:comparison} show that our method achieves the best performance. Scores for localization maps obtained using models trained in an unsupervised setting with 10k samples are reported. As the original paper of DMC \cite{deep_multimodal_clustering} only report scores for models trained with 400k samples, we use the scores of DMC trained on 10k samples reported in CAVL \cite{curriculum}.
\\[1ex]
\\[1ex]
\\[1ex]

\input{tab_ablation}
\input{fig_lim_vs_sup}
\section{Ablation Study}
We conducted ablation study to further analyze our proposed model. Specifically, we trained our model without the potential localization network and the selection module, and compared their performances. The altered model is obtained by fixing the potential localization map to be a constant ($P_{i, j} = const.$). The results are shown in Table \ref{tab:ablation}. As a result, the models without the potential localization network and the selection module worked as good as Senocak et al. \cite{learning_to_localize}, but worse than our proposed method, which means the potential localization network and the selection module gives a positive effect on the performance of the localization map. This result suggests that the potential localization network allows efficient training by Eq. \textcolor{blue}{(3)} because the workload of attention map $A_{i, j}$ is reduced by the potential localization map $P_{i, j}$, which predetermines the possible sound sources.

\section{Comparison between our unsupervised model and supervised model}
Visualization of potential localization maps for our unsupervised model and supervised model is shown in Fig. \ref{fig:lim_vs_sup}. It can be noted that for some samples, the potential localization map of the unsupervised model only responds to a small part of the observed object, whereas the supervised model mostly encompasess the entire object. This can be attributed to the fact that, as with existing unsupervised methods, our unsupervised model does not have any contraints that force the localization map and potential map to exhibit this trait. The model is only trained to minimize the similarity loss which does not necessarily require the model to respond to the entire object.

\section{Annotation Details}
The annotation process of the dataset used to analyze the performance gap in Table \textcolor{blue}{4} in the original paper is as follows. First, we randomly obtained 4K samples from Flickr-SoundNet \cite{soundnet} excluding the samples contained in the benchmark dataset \cite{learning_to_localize}. Then, we searched Type-B samples (there are several objects capable of producing sound in the image, but of which only one object is actually producing sound), by manually checking each image-sound pair. It should be noted that we had to listen to the sound in the annotation process to decide whether the samples are Type-A or Type-B, as their distinction is dependent on the accompanying sound as stated above. As a result, we found 30 Type-B samples in this process. We obtained the same number of Type-A samples to match the number of each sample. Finally, we annotated sound source for each sample using bounding boxes. In Fig. \ref{fig:typeA} and Fig. \ref{fig:typeB}, we show the examples of Type-A and Type-B images, and sound source annotations for them.

\section{Video}
The supplementary materials contain \textbf{``video.mp4''}. This video shows 4 cases where the localization map and the potential localization map are different. For instrument and machine sound, we use a 5-second clip from the original audio. For human sound, we use a 5-second clip from another video's sound that only contains human sound. For the visualization, we use the model trained with 144k samples and pretrained weights.

\section{Code}
The supplementary materials contain a jupyter notebook file \textbf{``code/code.ipynb''} and a html file \textbf{``code/code.html''}. The contents of these files are the same, but \textbf{``code/code.html''} does not require an environment for jupyter notebook. These files contain main parts of our code, including the network architectures and how we obtain the saliency maps. We also include \textbf{``code/requirements.txt''}, which shows the list of libraries we use.
\\[1ex]
\\[1ex]
\\[1ex]
\\[1ex]
\\[1ex]
\\
\\
\\
\\
\\
\\
\\
\\

\input{fig_typeA}
\input{fig_typeB}

%% file: tab_image_network.tex
\begin{table}[t]
    \caption{The architecture of the image network. ``conv'' indicates a 2-dimensional convolutional layer, ``size'' is the size of the filters, and ``n\_filters'' is the number of the filters. }
    \centering
    \scalebox{1}{
\begin{tabular}{cc}
Layer    & Layer information                                           \\ \hline
conv1    & (n\_filters = 64, size = 3, stride = 1, padding = 1), ReLU  \\
maxpool1 & (size = 2, stride = 2, padding = 0)                         \\
conv2    & (n\_filters = 128, size = 3, stride = 1, padding = 1), ReLU \\
maxpool2 & (size = 2, stride = 2, padding = 0)                         \\
conv3    & (n\_filters = 256, size = 3, stride = 1, padding = 1), ReLU \\
conv4    & (n\_filters = 256, size = 3, stride = 1, padding = 1), ReLU \\
maxpool3 & (size = 2, stride = 2, padding = 0)                         \\
conv5    & (n\_filters = 512, size = 3, stride = 1, padding = 1), ReLU \\
conv6    & (n\_filters = 512, size = 3, stride = 1, padding = 1), ReLU \\
maxpool4 & (size = 2, stride = 2, padding = 0)                         \\ \hline
\end{tabular}    }
    \label{tab:image_network}
\end{table}

%% file: tab_sound_network.tex
\begin{table}[t]
    \caption{The architecture of the sound network. ``BatchNorm'' means Batch Normalization \cite{batchnorm}.}
    \centering
    \scalebox{1}{
\begin{tabular}{cc}
Layer    & Layer information                                                      \\ \hline
conv1    & (n\_filters = 32, size = 3, stride = 1, padding = 1), BatchNorm, ReLU  \\
conv2    & (n\_filters = 32, size = 3, stride = 1, padding = 1), BatchNorm, ReLU  \\
maxpool1 & (size = 2, stride = 2, padding = 0)                                    \\
conv3    & (n\_filters = 64, size = 3, stride = 1, padding = 1), BatchNorm, ReLU  \\
conv4    & (n\_filters = 64, size = 3, stride = 1, padding = 1), BatchNorm, ReLU  \\
maxpool2 & (size = 2, stride = 2, padding = 0)                                    \\
conv5    & (n\_filters = 128, size = 3, stride = 1, padding = 1), BatchNorm, ReLU \\
conv6    & (n\_filters = 128, size = 3, stride = 1, padding = 1), BatchNorm, ReLU \\
maxpool3 & (size = 2, stride = 2, padding = 0)                                    \\
conv7    & (n\_filters = 128, size = 3, stride = 1, padding = 1), BatchNorm, ReLU \\
conv8    & (n\_filters = 128, size = 3, stride = 1, padding = 1), BatchNorm, ReLU \\
maxpool4 & (size = 2, stride = 2, padding = 0)                                    \\
conv9    & (n\_filters = 128, size = 3, stride = 1, padding = 1), BatchNorm, ReLU \\
conv10   & (n\_filters = 128, size = 3, stride = 1, padding = 1), BatchNorm, ReLU \\ \hline
\end{tabular}    }
    \label{tab:sound_network}
\end{table}

%% file: tab_N_saliency.tex
\begin{table}[t]
    \caption{Evaluation of saliency maps for top N classes for varying N. cIoU score with threshold 0.5 and AUC score are reported.}
    \centering
    \scalebox{1}{
\begin{tabular}{clcc}
\hline
\multirow{2}{*}{N} &  & \multicolumn{2}{c}{\begin{tabular}[c]{@{}c@{}}saliency \\ map\\ for top-N \\ classes\end{tabular}} \\ \cline{3-4} 
                   &  & cIoU                                             & AUC                                             \\ \hline
1                  &  & 45.7                                             & 41.1                                            \\
10                 &  & 51.5                                             & 45.1                                            \\
20                 &  & 52.1                                             & 45.7                                            \\
50                 &  & 52.6                                             & 46.2                                            \\
100                &  & 52.7                                             & 46.3                                            \\
200                &  & 52.5                                             & 46.1                                            \\
500                &  & 52.4                                             & 46.0                                            \\
1000               &  & 51.8                                             & 45.5                                            \\ \hline
\end{tabular}    }
    \label{tab:N_saliency}
\end{table}

%% file: tab_comparison.tex
\begin{table}[t]
    \caption{Additional comparison with other methods. Scores for localization maps obtained using models trained in an unsupervised setting with 10k samples are reported.}
    \centering
    \scalebox{1}{
\begin{tabular*}{7cm}{@{\extracolsep{\fill}}clclc}
\hline
\multicolumn{1}{c}{Method}               &  & \multicolumn{1}{c}{cIoU} &  & \multicolumn{1}{c}{AUC}  \\ \hline
\multicolumn{1}{c}{Ours}                 &  & \multicolumn{1}{c}{56.8} &  & \multicolumn{1}{c}{50.7} \\
\multicolumn{1}{c}{Senocak et al. {\cite{learning_to_localize}} (reported)} &  & \multicolumn{1}{c}{43.6} &  & \multicolumn{1}{c}{44.9} \\
DMC {\cite{deep_multimodal_clustering}} (reported)                 &  & 41.6                     &  & 45.2                     \\
CAVL {\cite{curriculum}} (reported)                &  & 50.0                     &  & 49.2                     \\
Two-stage {\cite{two_stage}} (reported)                     &  & 52.2                     &  & 49.6                     \\ \hline
\end{tabular*}    }
    \label{tab:comparison}
\end{table}

%% file: tab_ablation.tex
\begin{table*}[t]
    \caption{Evaluation of localization maps. cIoU score with threshold 0.5 and AUC score are reported. ``PLN, SM'' denotes the potential localization network and the selection module. All the experiments are conducted in an unsupervised setting.}
    \centering
    \scalebox{1}{
\begin{tabular*}{12cm}{@{\extracolsep{\fill}}cccccccccccccccc}
\hline
\begin{tabular}[c]{@{}c@{}}\# of \\ training\\ samples\end{tabular} &  & \multicolumn{2}{c}{\begin{tabular}[c]{@{}c@{}}\cite{learning_to_localize} \\ (reported)\end{tabular}} &  & \multicolumn{2}{c}{pretrained} &  & \multicolumn{2}{c}{\begin{tabular}[c]{@{}c@{}}pretrained \&\\ without  \\ PLN, SM\end{tabular}} &  & \multicolumn{2}{c}{de novo} &  & \multicolumn{2}{c}{\begin{tabular}[c]{@{}c@{}}de novo \& \\ without\\ PLN, SM\end{tabular}} \\ \cline{3-4} \cline{6-7} \cline{9-10} \cline{12-13} \cline{15-16} 
                                                                    &  & cIoU                                     & AUC                                      &  & cIoU           & AUC           &  & cIoU                                           & AUC                                            &  & cIoU         & AUC          &  & cIoU                                         & AUC                                          \\ \hline
1k                                                                  &  & ---                                      & ---                                      &  & 48.7           & 46.4          &  & 44.4                                           & 40.6                                           &  & 36.5         & 34.1         &  & 35.7                                         & 33.0                                         \\
2.5k                                                                &  & ---                                      & ---                                      &  & 50.3           & 47.7          &  & 48.7                                           & 45.8                                           &  & 40.7         & 37.3         &  & 38.6                                         & 36.6                                         \\
10k                                                                 &  & 43.6                                     & 44.9                                     &  & 56.8           & 50.7          &  & 56.4                                           & 49.2                                           &  & 48.4         & 45.3         &  & 47.6                                         & 44.3                                         \\
144k                                                                &  & 66.0                                     & 55.8                                     &  & 68.4           & 57.0          &  & 66.8                                           & 55.7                                           &  & 66.7         & 56.3         &  & 65.8                                         & 55.3                                         \\ \hline
\multirow{2}{*}{Random}                                             &  &                                          &                                          &  &                & \multicolumn{3}{c}{cIoU}                                          & \multicolumn{3}{c}{AUC}                                          &              &  &                                              &                                              \\ \cline{7-12}
                                                                    &  &                                          &                                          &  &                & \multicolumn{3}{c}{34.1}                                          & \multicolumn{3}{c}{32.3}                                         &              &  &                                              &                                              \\ \hline
\end{tabular*}    }
    \label{tab:ablation}
\end{table*}

%% file: fig_lim_vs_sup.tex
\begin{figure*}[h]
    \scalebox{0.83}{
        \begin{tabular}{p{7.3em} p{7.3em} p{7.3em} p{0.16em} p{7.3em} p{7.3em} p{7.3em}}
        %\toprule
        %-------
            \hfil Potential \hfil &
            \hfil Potential \hfil & 
            \hfil Ground truth \hfil &
            \hfil \hfil &
            \hfil Potential \hfil &
            \hfil Potential \hfil & 
            \hfil Ground truth \hfil 
            \\
                 \hfil Loc. map\hfil &
            \hfil Loc. map \hfil &
            \hfil \hfil  &
            \hfil \hfil &
            \hfil Loc. map\hfil &
            \hfil Loc. map\hfil &
            \hfil \hfil \\
            \hfil (Unsupervised) \hfil &
            \hfil (Supervised) \hfil &
            \hfil \hfil  &
            \hfil \hfil &
            \hfil (Unsupervised) \hfil &
            \hfil (Supervised) \hfil &
            \hfil \hfil \\
        %-------
        
            % 1st
            \begin{minipage}{7.3em}
                \centering
                \scalebox{0.42}
                {\includegraphics{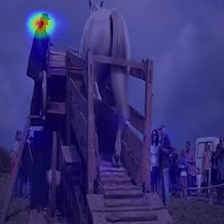}}
            \end{minipage} &
            % 2nd
            \begin{minipage}{7.3em}
                \centering
                \scalebox{0.42}
                {\includegraphics{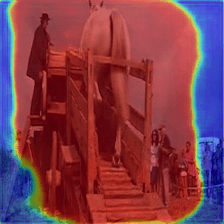}}
            \end{minipage} &
            % 3rd
            \begin{minipage}{7.3em}
                \centering
                \scalebox{0.42}
                {\includegraphics{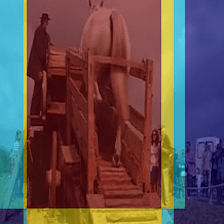}}
            \end{minipage} &
            \begin{minipage}{0.16em}
            \end{minipage} &
            % 4th
            \begin{minipage}{7.3em}
                \centering
                \scalebox{0.42}
                {\includegraphics{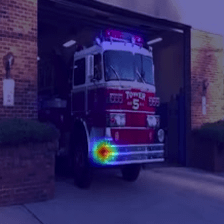}}
            \end{minipage} &
            
            \begin{minipage}{7.3em}
                \centering
                \scalebox{0.42}
                {\includegraphics{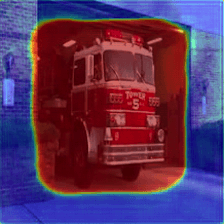}}
            \end{minipage} &
            % 4th
            \begin{minipage}{7.3em}
                \centering
                \scalebox{0.42}
                {\includegraphics{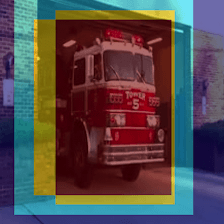}}
            \end{minipage} \vspace{0.001mm} \\
            %-------
        %-------
        %-------
            % 1st
            \begin{minipage}{7.3em}
                \centering
                \scalebox{0.42}
                {\includegraphics{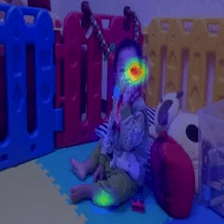}}
            \end{minipage} &
            % 2nd
            \begin{minipage}{7.3em}
                \centering
                \scalebox{0.42}
                {\includegraphics{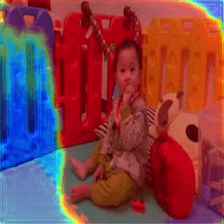}}
            \end{minipage} &
            % 3rd
            \begin{minipage}{7.3em}
                \centering
                \scalebox{0.42}
                {\includegraphics{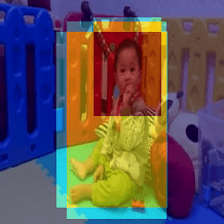}}
            \end{minipage} &
            \begin{minipage}{0.16em}
            \end{minipage} &
            % 4th
            \begin{minipage}{7.3em}
                \centering
                \scalebox{0.42}
                {\includegraphics{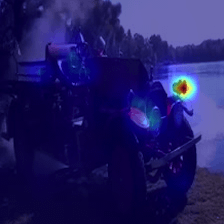}}
            \end{minipage} &
            
            \begin{minipage}{7.3em}
                \centering
                \scalebox{0.42}
                {\includegraphics{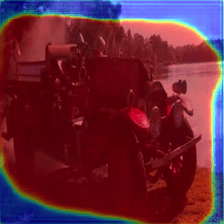}}
            \end{minipage} &
            % 4th
            \begin{minipage}{7.3em}
                \centering
                \scalebox{0.42}
                {\includegraphics{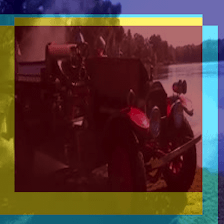}}
            \end{minipage} \vspace{0.001mm} \\
        %-------
        %-------
            %-------
            % 1st
            \begin{minipage}{7.3em}
                \centering
                \scalebox{0.42}
                {\includegraphics{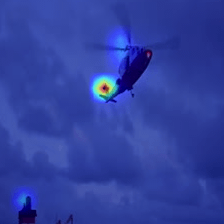}}
            \end{minipage} &
            % 2nd
            \begin{minipage}{7.3em}
                \centering
                \scalebox{0.42}
                {\includegraphics{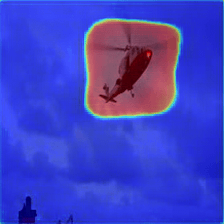}}
            \end{minipage} &
            % 3rd
            \begin{minipage}{7.3em}
                \centering
                \scalebox{0.42}
                {\includegraphics{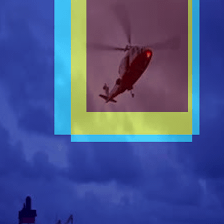}}
            \end{minipage} &
            \begin{minipage}{0.16em}
            \end{minipage} &
            % 4th
            \begin{minipage}{7.3em}
                \centering
                \scalebox{0.42}
                {\includegraphics{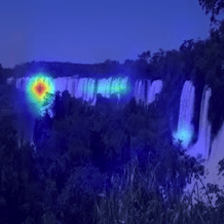}}
            \end{minipage} &
            
            \begin{minipage}{7.3em}
                \centering
                \scalebox{0.42}
                {\includegraphics{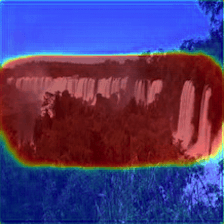}}
            \end{minipage} &
            % 4th
            \begin{minipage}{7.3em}
                \centering
                \scalebox{0.42}
                {\includegraphics{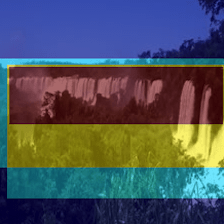}}
            \end{minipage} \\
            %-------
        
        %\bottomrule
        %-------
        \end{tabular}     }
    	\caption{Visualization of potential localization maps for our unsupervised model and supervised model. For the unsupervised setting, we used the model trained with 144k samples and pretrained weights.}
    \label{fig:lim_vs_sup} % \ref{label}で表番号を参照
\end{figure*}

%% file: fig_typeA.tex
\begin{figure*}[t]
    \scalebox{0.59}{
        \begin{tabular}{p{7.6em} p{7.6em} p{7.6em} p{7.6em} p{0.16em} p{7.6em} p{7.6em} p{7.6em} p{7.6em}}
        %\toprule
        %-------
            \hfil Original Image\hfil &
            \hfil Ground truth\hfil & 
            \hfil Potential \hfil &
            \hfil Loc. map\hfil &
            \hfil \hfil &
            \hfil Original Image\hfil &
            \hfil Ground truth\hfil & 
            \hfil Potential \hfil &
            \hfil Loc. map\hfil 
            \\
            \hfil \hfil &
            \hfil \hfil & 
            \hfil Loc. map\hfil &
            \hfil \hfil &
            \hfil \hfil &
            \hfil \hfil &
            \hfil \hfil & 
            \hfil Loc. map\hfil &
            \hfil \hfil 
            \\
            \begin{minipage}{7.6em}
                \centering
                \scalebox{0.435}
                {\includegraphics{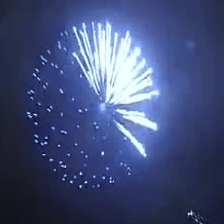}}
            \end{minipage} &
            \begin{minipage}{7.6em}
                \centering
                \scalebox{0.435}
                {\includegraphics{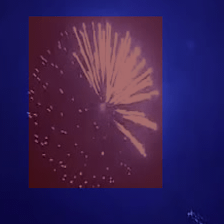}}
            \end{minipage} &
            % 2nd
            \begin{minipage}{7.6em}
                \centering
                \scalebox{0.435}
                {\includegraphics{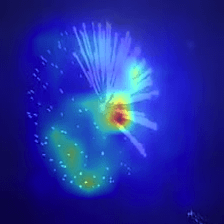}}
            \end{minipage} &
            % 3rd
            \begin{minipage}{7.6em}
                \centering
                \scalebox{0.435}
                {\includegraphics{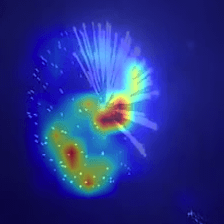}}
            \end{minipage} &
            \begin{minipage}{0.16em}
            \end{minipage} &
            % 4th
            \begin{minipage}{7.6em}
                \centering
                \scalebox{0.435}
                {\includegraphics{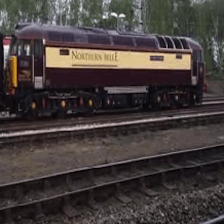}}
            \end{minipage} &
            \begin{minipage}{7.6em}
                \centering
                \scalebox{0.435}
                {\includegraphics{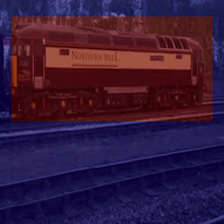}}
            \end{minipage} &
            \begin{minipage}{7.6em}
                \centering
                \scalebox{0.435}
                {\includegraphics{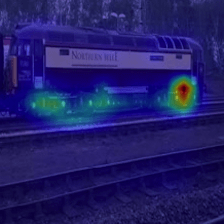}}
            \end{minipage} &
            % 4th
            \begin{minipage}{7.6em}
                \centering
                \scalebox{0.435}
                {\includegraphics{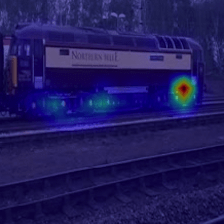}}
            \end{minipage} \vspace{0.01mm} \\
            %-------
            \begin{minipage}{7.6em}
                \centering
                \scalebox{0.435}
                {\includegraphics{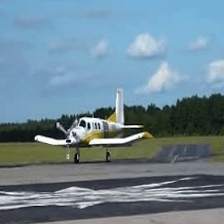}}
            \end{minipage} &
            \begin{minipage}{7.6em}
                \centering
                \scalebox{0.435}
                {\includegraphics{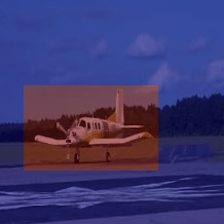}}
            \end{minipage} &
            % 2nd
            \begin{minipage}{7.6em}
                \centering
                \scalebox{0.435}
                {\includegraphics{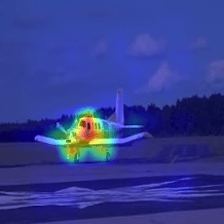}}
            \end{minipage} &
            % 3rd
            \begin{minipage}{7.6em}
                \centering
                \scalebox{0.435}
                {\includegraphics{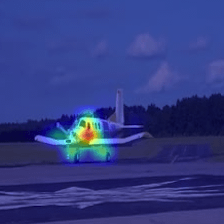}}
            \end{minipage} &
            \begin{minipage}{0.16em}
            \end{minipage} &
            % 4th
            \begin{minipage}{7.6em}
                \centering
                \scalebox{0.435}
                {\includegraphics{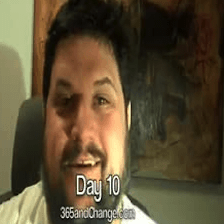}}
            \end{minipage} &
            \begin{minipage}{7.6em}
                \centering
                \scalebox{0.435}
                {\includegraphics{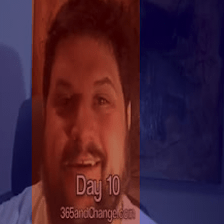}}
            \end{minipage} &
            \begin{minipage}{7.6em}
                \centering
                \scalebox{0.435}
                {\includegraphics{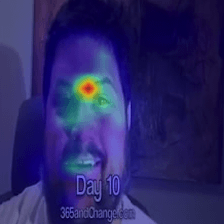}}
            \end{minipage} &
            % 4th
            \begin{minipage}{7.6em}
                \centering
                \scalebox{0.435}
                {\includegraphics{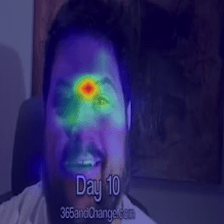}}
            \end{minipage} \vspace{0.1mm} \\
            %-------
        %\bottomrule
        %-------
        \end{tabular}     }
    	\caption{Visualization of potential localization maps and localization maps of Type A along with the ground truth. For the visualization, we used the model trained with 144k samples and pretrained weights.}
    \label{fig:typeA} % \ref{label}で表番号を参照
\end{figure*}

%% file: fig_typeB.tex
\begin{figure*}[t]
    \scalebox{0.59}{
        \begin{tabular}{p{7.6em} p{7.6em} p{7.6em} p{7.6em} p{0.16em} p{7.6em} p{7.6em} p{7.6em} p{7.6em}}
        %\toprule
        %-------
            \hfil Original Image\hfil &
            \hfil Ground truth\hfil & 
            \hfil Potential \hfil &
            \hfil Loc. map\hfil &
            \hfil \hfil &
            \hfil Original Image\hfil &
            \hfil Ground truth\hfil & 
            \hfil Potential \hfil &
            \hfil Loc. map\hfil 
            \\
            \hfil \hfil &
            \hfil \hfil & 
            \hfil Loc. map\hfil &
            \hfil \hfil &
            \hfil \hfil &
            \hfil \hfil &
            \hfil \hfil & 
            \hfil Loc. map\hfil &
            \hfil \hfil 
            \\
            \begin{minipage}{7.6em}
                \centering
                \scalebox{0.435}
                {\includegraphics{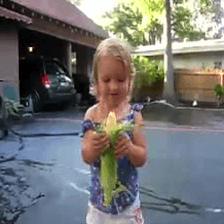}}
            \end{minipage} &
            \begin{minipage}{7.6em}
                \centering
                \scalebox{0.435}
                {\includegraphics{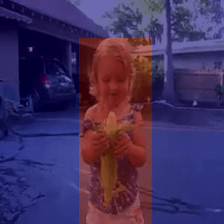}}
            \end{minipage} &
            % 2nd
            \begin{minipage}{7.6em}
                \centering
                \scalebox{0.435}
                {\includegraphics{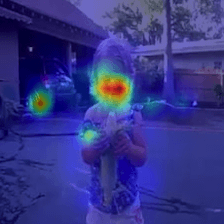}}
            \end{minipage} &
            % 3rd
            \begin{minipage}{7.6em}
                \centering
                \scalebox{0.435}
                {\includegraphics{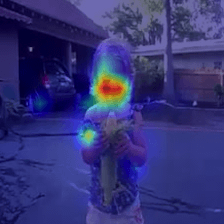}}
            \end{minipage} &
            \begin{minipage}{0.16em}
            \end{minipage} &
            % 4th
            \begin{minipage}{7.6em}
                \centering
                \scalebox{0.435}
                {\includegraphics{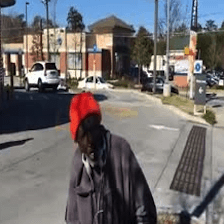}}
            \end{minipage} &
            \begin{minipage}{7.6em}
                \centering
                \scalebox{0.435}
                {\includegraphics{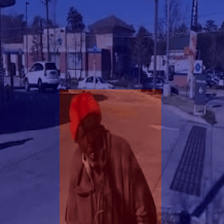}}
            \end{minipage} &
            \begin{minipage}{7.6em}
                \centering
                \scalebox{0.435}
                {\includegraphics{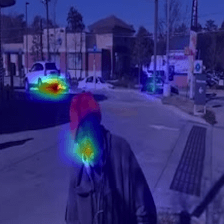}}
            \end{minipage} &
            % 4th
            \begin{minipage}{7.6em}
                \centering
                \scalebox{0.435}
                {\includegraphics{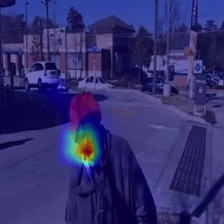}}
            \end{minipage} \vspace{0.01mm} \\
            %-------
            \begin{minipage}{7.6em}
                \centering
                \scalebox{0.435}
                {\includegraphics{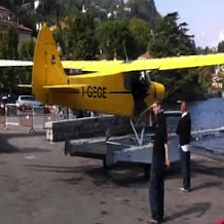}}
            \end{minipage} &
            \begin{minipage}{7.6em}
                \centering
                \scalebox{0.435}
                {\includegraphics{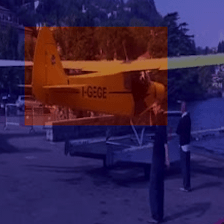}}
            \end{minipage} &
            % 2nd
            \begin{minipage}{7.6em}
                \centering
                \scalebox{0.435}
                {\includegraphics{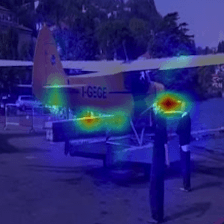}}
            \end{minipage} &
            % 3rd
            \begin{minipage}{7.6em}
                \centering
                \scalebox{0.435}
                {\includegraphics{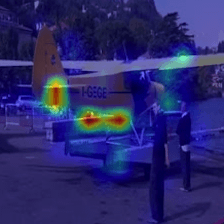}}
            \end{minipage} &
            \begin{minipage}{0.16em}
            \end{minipage} &
            % 4th
            \begin{minipage}{7.6em}
                \centering
                \scalebox{0.435}
                {\includegraphics{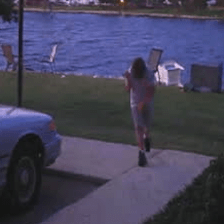}}
            \end{minipage} &
            \begin{minipage}{7.6em}
                \centering
                \scalebox{0.435}
                {\includegraphics{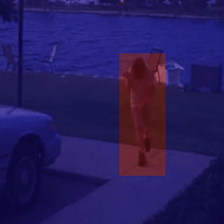}}
            \end{minipage} &
            \begin{minipage}{7.6em}
                \centering
                \scalebox{0.435}
                {\includegraphics{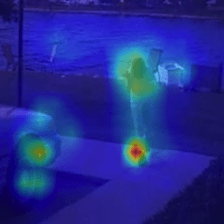}}
            \end{minipage} &
            % 4th
            \begin{minipage}{7.6em}
                \centering
                \scalebox{0.435}
                {\includegraphics{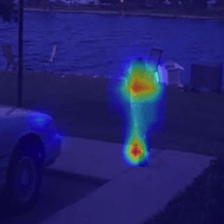}}
            \end{minipage} \vspace{0.1mm} \\
            %-------
        %\bottomrule
        %-------
        \end{tabular}     }
    	\caption{Visualization of potential localization maps and localization maps of Type B along with the ground truth. For the visualization, we used the model trained with 144k samples and pretrained weights.}
    \label{fig:typeB}
\end{figure*}

%% file: main.bbl
\begin{thebibliography}{10}

\bibitem{randomforest}
Breiman, L.:
\newblock Random forests.
\newblock Machine Learning \textbf{45} (2001)  5--32

\bibitem{xgboost}
Chen, T., Guestrin, C.:
\newblock Xgboost: A scalable tree boosting system.
\newblock In: International Conference on Knowledge Discovery and Data Mining
  (KDD). (2016)

\bibitem{lightgbm}
Ke, G., Meng, Q., Finley, T., Wang, T., Chen, W., Ma, W., Ye, Q., Liu, T.Y.:
\newblock Lightgbm: A highly efficient gradient boosting decision tree.
\newblock In: Neural Information Processing Systems (NIPS). (2017)

\bibitem{catboost}
Prokhorenkova, L., Gusev, G., Vorobev, A., Dorogush, A.V., Gulin, A.:
\newblock Catboost: Unbiased boosting with categorical features.
\newblock In: Neural Information Processing Systems (NIPS). (2018)

\bibitem{LLL}
Arandjelovic, R., Zisserman, A.:
\newblock Look, listen and learn.
\newblock In: International Conference on Computer Vision (ICCV). (2017)

\bibitem{objects_that_sound}
Arandjelovi{\'c}, R., Zisserman, A.:
\newblock Objects that sound.
\newblock In: European Conference on Computer Vision (ECCV). (2018)

\bibitem{deep_multimodal_clustering}
Hu, D., Nie, F., Li, X.:
\newblock Deep multimodal clustering for unsupervised audiovisual learning.
\newblock In: Computer Vision and Pattern Recognition (CVPR). (2019)

\bibitem{co_segmentation}
Rouditchenko, A., Zhao, H., Gan, C., McDermott, J., Torralba, A.:
\newblock Self-supervised audio-visual co-segmentation.
\newblock In: International Conference on Acoustics, Speech and Signal
  Processing (ICASSP). (2019)

\bibitem{sound_of_pixels}
Zhao, H., Gan, C., Rouditchenko, A., Vondrick, C., McDermott, J., Torralba, A.:
\newblock The sound of pixels.
\newblock In: European Conference on Computer Vision (ECCV). (2018)

\bibitem{multisensory}
Owens, A., Efros, A.A.:
\newblock Audio-visual scene analysis with self-supervised multisensory
  features.
\newblock In: European Conference on Computer Vision (ECCV). (2018)

\bibitem{learning_to_localize}
Senocak, A., Oh, T.H., Kim, J., Yang, M.H., So~Kweon, I.:
\newblock Learning to localize sound source in visual scenes.
\newblock In: Computer Vision and Pattern Recognition (CVPR). (2018)

\bibitem{sound_of_motions}
Zhao, H., Gan, C., Ma, W.C., Torralba, A.:
\newblock The sound of motions.
\newblock In: International Conference on Computer Vision (ICCV). (2019)

\bibitem{see_hear_read}
Aytar, Y., Vondrick, C., Torralba, A.:
\newblock See, hear, and read: Deep aligned representations.
\newblock arXiv preprint arXiv:1706.00932 (2017)

\bibitem{soundnet}
Aytar, Y., Vondrick, C., Torralba, A.:
\newblock Soundnet: Learning sound representations from unlabeled video.
\newblock In: Neural Information Processing Systems (NIPS). (2016)

\bibitem{word_unit}
Harwath, D., Glass, J.R.:
\newblock Learning word-like units from joint audio-visual analysis.
\newblock In: Association for Computational Linguistics (ACL). (2017)

\bibitem{spoken_language}
Harwath, D., Torralba, A., Glass, J.:
\newblock Unsupervised learning of spoken language with visual context.
\newblock In: Neural Information Processing Systems (NIPS). (2016)

\bibitem{ambient_sound}
Owens, A., Wu, J., McDermott, J.H., Freeman, W.T., Torralba, A.:
\newblock Ambient sound provides supervision for visual learning.
\newblock In: European Conference on Computer Vision (ECCV). (2016)

\bibitem{event_localization}
Tian, Y., Shi, J., Li, B., Duan, Z., Xu, C.:
\newblock Audio-visual event localization in unconstrained videos.
\newblock In: European Conference on Computer Vision (ECCV). (2018)

\bibitem{cooperative_learning}
Korbar, B., Tran, D., Torresani, L.:
\newblock Cooperative learning of audio and video models from self-supervised
  synchronization.
\newblock In: Neural Information Processing Systems (NIPS). (2018)

\bibitem{blind_old}
Lyon, R.F.:
\newblock A computational model of binaural localization and separation.
\newblock In: International Conference on Acoustics, Speech and Signal
  Processing (ICASSP). (1983)

\bibitem{blind1}
Hershey, J.R., Chen, Z., Roux, J.L., Watanabe, S.:
\newblock Deep clustering: Discriminative embeddings for segmentation and
  separation.
\newblock In: International Conference on Acoustics, Speech and Signal
  Processing (ICASSP). (2016)

\bibitem{blind2}
Chen, Z., Luo, Y., Mesgarani, N.:
\newblock Deep attractor network for single-microphone speaker separation.
\newblock In: International Conference on Acoustics, Speech and Signal
  Processing (ICASSP). (2017)

\bibitem{blind3}
Yu, D., Kolb{\ae}k, M., Tan, Z.H., Jensen, J.:
\newblock Permutation invariant training of deep models for speaker-independent
  multi-talker speech separation.
\newblock In: International Conference on Acoustics, Speech and Signal
  Processing (ICASSP). (2017)

\bibitem{motion_informed}
Parekh, S., Essid, S., Ozerov, A., Duong, N., P{\'e}rez, P., Richard, G.:
\newblock Motion informed audio source separation.
\newblock In: International Conference on Acoustics, Speech and Signal
  Processing (ICASSP). (2017)

\bibitem{two_multimodal}
Sedighin, F., {Babaie-Zadeh}, M., Rivet, B., Jutten, C.:
\newblock Two multimodal approaches for single microphone source separation.
\newblock In: European Signal Processing Conference (EUSIPCO). (2016)

\bibitem{learning_to_separate}
Gao, R., Feris, R., Grauman, K.:
\newblock Learning to separate object sounds by watching unlabeled video.
\newblock In: European Conference on Computer Vision (ECCV). (2018)

\bibitem{independent_components}
Smaragdis, P., Casey, M.:
\newblock Audio/visual independent components.
\newblock In: International Conference on Independent Component Analysis and
  Signal Separation (ICA). (2003)

\bibitem{sparsity}
Pu, J., Panagakis, Y., Petridis, S., Pantic, M.:
\newblock Audio-visual object localization and separation using low-rank and
  sparsity.
\newblock In: International Conference on Acoustics, Speech and Signal
  Processing (ICASSP). (2017)

\bibitem{co_separating}
Gao, R., Grauman, K.:
\newblock Co-separating sounds of visual objects.
\newblock In: International Conference on Computer Vision (ICCV). (2019)

\bibitem{cocktail}
Ephrat, A., Mosseri, I., Lang, O., Dekel, T., Wilson, K., Hassidim, A.,
  Freeman, W.T., Rubinstein, M.:
\newblock Looking to listen at the cocktail party: A speaker-independent
  audio-visual model for speech separation.
\newblock In: Special Interest Group on Computer GRAPHics and Interactive
  Techniques (SIGGRAPH). (2018)

\bibitem{visual_speech_enhancement}
Gabbay, A., Shamir, A., Peleg, S.:
\newblock Visual speech enhancement.
\newblock In: International Speech Communication Association (INTERSPEECH).
  (2018)

\bibitem{conversation_enhancement}
Afouras, T., Chung, J.S., Zisserman, A.:
\newblock The conversation: Deep audio-visual speech enhancement.
\newblock In: International Speech Communication Association (INTERSPEECH).
  (2018)

\bibitem{blind_audiovisual_source_separation}
Llagostera~Casanovas, A., Monaci, G., Vandergheynst, P., Gribonval, R.:
\newblock Blind audiovisual source separation based on sparse redundant
  representations.
\newblock Transactions on Multimedia \textbf{12} (2010)  358--371

\bibitem{robotics2}
Nakadai, K., Okuno, H.G., Kitano, H.:
\newblock Real-time sound source localization and separation for robot
  audition.
\newblock In: International Conference on Spoken Language Processing (ICSLP).
  (2002)

\bibitem{robotics}
Argentieri, S., Dan{\`e}s, P., Sou{\`e}res, P.:
\newblock A survey on sound source localization in robotics: From binaural to
  array processing methods.
\newblock Comput. Speech Language \textbf{34} (2015)  87--112

\bibitem{robotics3}
Nakamura, K., Nakadai, K., Asano, F., Ince, G.:
\newblock Intelligent sound source localization and its application to
  multimodal human tracking.
\newblock In: International Conference on Intelligent Robots and Systems
  (IROS). (2011)

\bibitem{robotics4}
Strobel, N., Spors, S., Rabenstein, R.:
\newblock Joint audio-video object localization and tracking.
\newblock Signal Processing Magazine \textbf{18} (2001)  22--31

\bibitem{synchrony}
Hershey, J.R., Movellan, J.R.:
\newblock Audio vision: Using audio-visual synchrony to locate sounds.
\newblock In: Neural Information Processing Systems (NIPS). (1999)

\bibitem{pixels_that_sound}
Kidron, E., Schechner, Y.Y., Elad, M.:
\newblock Pixels that sound.
\newblock In: Computer Vision and Pattern Recognition (CVPR). (2005)

\bibitem{statistical_models}
Fisher~III, J.W., Darrell, T., Freeman, W.T., Viola, P.A.:
\newblock Learning joint statistical models for audio-visual fusion and
  segregation.
\newblock In: Neural Information Processing Systems (NIPS). (2001)

\bibitem{harmony_in_motion}
Barzelay, Z., Schechner, Y.:
\newblock Harmony in motion.
\newblock In: Computer Vision and Pattern Recognition (CVPR). (2007)

\bibitem{alexnet}
Krizhevsky, A., Sutskever, I., Hinton, G.E.:
\newblock Imagenet classification with deep convolutional neural networks.
\newblock In: Neural Information Processing Systems (NIPS). (2012)

\bibitem{VGG}
Simonyan, K., Zisserman, A.:
\newblock Very deep convolutional networks for large-scale image recognition.
\newblock In: International Conference on Learning Representations (ICLR).
  (2015)

\bibitem{imagenet}
Deng, J., Dong, W., Socher, R., Li, L.J., Li, K., Fei-Fei, L.:
\newblock Imagenet: A large-scale hierarchical image database.
\newblock In: Computer Vision and Pattern Recognition (CVPR). (2009)

\bibitem{ReLU}
Nair, V., Hinton, G.E.:
\newblock Rectified linear units improve restricted boltzmann machines.
\newblock In: International Conference on Machine Learning (ICML). (2010)

\bibitem{NIN}
Lin, M., Chen, Q., Yan, S.:
\newblock Network in network.
\newblock In: International Conference on Learning Representations (ICLR).
  (2014)

\bibitem{siamese}
Bromley, J., Guyon, I., LeCun, Y., S{\"a}ckinger, E., Shah, R.:
\newblock Signature verification using a "siamese" time delay neural network.
\newblock In: Neural Information Processing Systems (NIPS). (1994)

\bibitem{grad_cam}
Selvaraju, R.R., Cogswell, M., Das, A., Vedantam, R., Parikh, D., Batra, D.:
\newblock Grad-cam: Visual explanations from deep networks via gradient-based
  localization.
\newblock In: International Conference on Computer Vision (ICCV). (2017)

\bibitem{UNet}
Ronneberger, O., Fischer, P., Brox, T.:
\newblock U-net: Convolutional networks for biomedical image segmentation.
\newblock In: Medical Image Computing and Computer-Assisted Intervention
  (MICCAI). (2015)

\bibitem{adam}
Kingma, D.P., Ba, J.:
\newblock Adam: A method for stochastic optimization.
\newblock In: International Conference on Learning Representations (ICLR).
  (2015)

\bibitem{pytorch}
Paszke, A., Gross, S., Chintala, S., Chanan, G., Yang, E., DeVito, Z., Lin, Z.,
  Desmaison, A., Antiga, L., Lerer, A.:
\newblock Automatic differentiation in pytorch.
\newblock (2017)

\bibitem{resnet}
He, K., Zhang, X., Ren, S., Sun, J.:
\newblock Deep residual learning for image recognition.
\newblock In: Computer Vision and Pattern Recognition (CVPR). (2016)

\bibitem{batchnorm}
Ioffe, S., Szegedy, C.:
\newblock Batch normalization: Accelerating deep network training by reducing
  internal covariate shift.
\newblock In: International Conference on Machine Learning (ICML). (2015)

\bibitem{curriculum}
Hu, D., Wang, Z., Xiong, H., Wang, D., Nie, F., Dou, D.:
\newblock Curriculum audiovisual learning.
\newblock arXiv preprint arXiv:2001.09414 (2020)

\bibitem{two_stage}
Qian, R., Hu, D., Dinkel, H., Wu, M., Xu, N., Lin, W.:
\newblock A two-stage framework for multiple sound-source localization.
\newblock In: Computer Vision and Pattern Recognition Workshops (CVPRW). (2020)

\end{thebibliography}
